\documentclass[journal]{IEEEtran}

\usepackage{lineno,hyperref}
\usepackage{bm}
\usepackage{mathrsfs}
\usepackage{amsmath}
\usepackage{amssymb}
\usepackage{graphicx}
\usepackage{subfigure}
\usepackage{epstopdf}
\usepackage{xcolor}
\usepackage{wasysym}
\usepackage{footnote}
\usepackage{amssymb}
\usepackage{textcomp}
\usepackage{overpic}
\modulolinenumbers[5]

\ifCLASSINFOpdf

\else

\fi

\begin{document}

\title{Study of Diffusion Normalized Least Mean M-estimate Algorithms}

\author{Yi~Yu,~\IEEEmembership{Member,~IEEE},
        ~Hongsen He,~\IEEEmembership{Member,~IEEE},
        ~Tao Yang,
        ~Xueyuan Wang,
        ~Rodrigo C.~de Lamare,~\IEEEmembership{Senior Member,~IEEE}
\thanks{This work was partially supported by the National Natural Science Foundation of China (NSFC) (Nos. 61901400, 61571376, and 61771411), and the Doctoral Research Fund of Southwest University of Science and Technology in China (No. 19zx7122).}

\thanks{Y. Yu, H. He, T. Yang, and X. Wang are with School of Information Engineering, Robot Technology Used for Special Environment Key Laboratory of Sichuan Province, Southwest University of Science and Technology, Mianyang, 621010, China (e-mail: yuyi\_xyuan@163.com, hongsenhe@gmail.com, yangtao@swust.edu.cn, 121053406@qq.com).}
\thanks{R. C. de Lamare is with CETUC, PUC-Rio, Rio de Janeiro 22451-900, Brazil, and also with the Department of Electronics, University of York, York YO10 5DD, U.K. (e-mail: rcdl500@ohm.york.ac.uk).}
}


\maketitle

\begin{abstract}
This work proposes diffusion normalized least mean M-estimate algorithm based on the modified Huber function, which can equip distributed networks with robust learning capability in the presence of impulsive interference. In order to exploit the system's underlying sparsity to further improve the learning performance, a sparse-aware variant is also developed by incorporating the $l_0$-norm of the estimates into the update process. We then analyze the transient, steady-state and stability behaviors of the algorithms in a unified framework. In particular, we present an analytical method that is simpler than conventional approaches to deal with the score function since it removes the requirements of integrals and Price's theorem. Simulations in various impulsive noise scenarios show that the proposed algorithms are superior to some existing diffusion algorithms and the theoretical results are verifiable.
\end{abstract}


\IEEEpeerreviewmaketitle

\section{Introduction}

\IEEEPARstart{I}{n} the last decade, distributed adaptive estimation
over networks with numerous sensing agents/nodes has been studied
frequently~\cite{sayed2014adaptation,sayed2013diffusion}, and
applied to many areas such as spectrum estimation in cognitive
radios~\cite{di2013distributed,miller2016distributed,7458257} and
beamforming~\cite{6292866}. In distributed adaptive estimation, the
nodes cooperate with each other through the network links, to
estimate parameters of interest using the streaming measurements. In
the light of different cooperation strategies among interconnected
nodes, distributed adaptive algorithms can be categorized into
incremental~\cite{LLi2010},
consensus~\cite{4787093,theodoridis2015machine,tu2012diffusion,wang2019distributed},
and
diffusion~\cite{lopes2008diffusion,xu2015distributed,8306496,cattivelli2010diffusion,chen2012diffusion,chen2013distributed}
strategies.

The incremental strategy requires a cyclic communication path that
covers all nodes, which is prone to link and node
failures~\cite{LLi2010}. For both consensus and diffusion
strategies, the nodes exchange local information with their
neighboring nodes without the need for such a cycle. The consensus
strategy enforces an agreement constraint among the nodes. To solve
the global mean-square-error (MSE) minimization problem under this
constraint, the alternating-direction method of multipliers
(ADMM)~\cite{4787093} and the stochastic gradient descent
(SGD)~\cite[Section 5.13.4]{theodoridis2015machine} methods were
used and different distributed consensus algorithms were presented.
The work in~\cite{tu2012diffusion} has proved that the stability of
the SGD-based consensus algorithm is dependent of the network
topology, while the diffusion algorithm does not have this
limitation and has better performance. It has been shown
in~\cite{cattivelli2010diffusion} that the diffusion algorithm
slightly outperforms the ADMM-based consensus algorithm in terms of
the steady-state performance for ideal communication links. So, the
focus of this work is on diffusion-based distributed algorithms,
among which the adapt-then-combine (ATC) implementation is of
particular interest, since it usually outperform the
combine-then-adapt (CTA)
implementation~\cite{cattivelli2010diffusion,sayed2014adaptation}.
Moreover, the CTA implementation can also be obtained
straightforwardly from the ATC one. For brevity, we would leave out
'ATC' in the following texts.

In signal processing, the measurement noise is usually assumed to be
Gaussian. In this case, the MSE criterion is widely used for
developing many algorithms such as diffusion least mean-square
(DLMS)~\cite{lopes2008diffusion} and normalized
DLMS~(DNLMS)~\cite{jung2015variable}. In real-world applications,
however, the measurements may be corrupted by the non-Gaussian noise
including Gaussian and impulsive components. Its probability density
function (pdf) has heavier tail than that of Gaussian noise. Such
noise may be natural or from man-made, e.g., biological noise, pulse
electromagnetic interference, and keyboard clicking or pen dropping
in
teleconferences~\cite{blackard1993measurements,georgiou1999alpha,chitre2006optimal,chen2017robust,chen2018mixture,Yu2019}.
Although the realization of impulsive noise in the time domain is
sparse, its amplitude is much higher than that of the nominal
measurement; thus, it severely leads to the performance degeneration
of the above algorithms, or even divergence. Especially due to the
cooperation between nodes, impulsive noise that occurs even at one
of the nodes could be propagated over the entire network.

In order to obtain good estimation performance over networks
disturbed by impulsive noise, several robust diffusion algorithms
have been
proposed~\cite{aifir,blackard1993measurements,jio,rrsgp,spa,smtvb,jidf,jidf_echo,sjidf,ccg,jiocdma,jiomimo,tds,mbdf,rrstap,l1stap,l1stap2,rccm,dfjio,locsme,rrser,rdrcb,rrdoa,okspme,kaesprit},
\cite{wen2013diffusion,ni2016diffusion,al2017robust,chen2018diffusion,kumar2015robust,li2018diffusion}.
Specifically, some examples based on different robust minimization
criteria are the diffusion least mean $p$-th error (DLMP)
where~$1\leq p<2$~\cite{wen2013diffusion}, diffusion sign error LMS
(DSE-LMS)~\cite{ni2016diffusion}, diffusion error-nonlinearity LMS
(DEN-LMS)~\cite{al2017robust}, and diffusion least logarithmic
absolute difference (D-LLAD)~\cite{chen2018diffusion} algorithms.
However, the robustness of DLMP against impulsive noise depends on
the value of $p$ and the parameters in the $\alpha$-stable~impulsive
noise. The DSE-LMS is a particular case of the DLMP when $p=1$,
which is usually a good performance benchmark as compared to other
algorithms in impulsive noise. As shown in \cite{al2017robust}, the
DEN-LMS converges more slowly than the DSE-LMS in Laplacian noise
environments. The D-LLAD achieves faster convergence than the
DSE-LMS~\cite{chen2018diffusion}. Owing to the capability of Huber
M-estimate function for removing outliers,
references~\cite{kumar2015robust} and~\cite{li2018diffusion}
proposed the diffusion Huber LMS and NLMS algorithms, respectively,
and their estimation performance in impulsive noise relies on an
empirical threshold. Also, the theoretical behaviors of both
Huber-based algorithms have not been studied.

On the other hand, the parameter vector of interest may be sparse, which means only a fraction of elements in the parameter vector are relatively large and the remaining coefficients are small enough to be negligible or zero. Such sparsity is often encountered in many situations; to name just a few, spectrum estimation~\cite{di2013distributed,miller2016distributed}, compressed sensing~\cite{jin2010stochastic}, and Digital TV transmission channels~\cite{schreiber1995advanced}. Thus, exploiting the underlying sparsity is able to improve the estimation performance. A cost-effective approach is to add a sparse regularization constraint (e.g., the $l_1$-norm or the $l_0$-norm) on the estimated vector into the cost function, known as the sparsity-aware technique~\cite{yu2018sparsity,gu2009l,olinto2016transient,wu2013gradient,lima2013affine}. Recently, the merit of sparsity has also been extended to distributed estimation, and several sparsity-aware diffusion algorithms were developed~\cite{liu2012diffusion,di2013sparse,ShravanPoly2018,liu2014distributed,zheng2017diffusion}. Nevertheless, it should be remarked that distributed estimation with sparsity in impulsive noise has not yet drawn much attention.

This work focuses on developing and analyzing the M-estimate based diffusion algorithms over distributed networks in the presence of impulsive noise. The main contributions are:

1) We propose a novel diffusion normalized least mean M-estimate (D-NLMM) algorithm by applying the modified Huber (MH) function with adaptive thresholds, which is robust against impulsive noise. For scenarios with sparse parameter vectors, we further develop the sparsity-aware D-NLMM (D-SNLMM) algorithm by incorporating the $l_0$-norm based regularization technique.

2) Based on the contaminated-Gaussian (CG) noise model, the mean and mean-square behaviors of the proposed algorithms are analyzed and then supported by simulations. These analyses are addressed in a unified manner from the D-SNLMM update. In particular, we provide a closed-form expression for predicting the steady-state performance of the D-NLMM algorithm. The analysis results show that the stability conditions for both D-NLMM and D-SNLMM algorithms are independent of the powers of input regressors.

3) We also present proximal variants of the D-SNLMM algorithm which exploits the sparsity by virtue of the forward-backward splitting method.

4) Simulation results in various noise environments demonstrate the superiority of the proposed algorithms.

The paper is organized as follows. Section~II introduces the signal model and the DNLMS algorithm. In Section~III, we present the derivations of the D-NLMM and the D-SNLMM algorithms. In Section~IV, we carry out a stochastic analysis of the proposed algorithms. Simulation results are given in Section~V. Section~VI draws some conclusions.

\emph{Notations:} We use the subscript on the time index $i$ to denote matrices and vectors, and the parentheses on the time index $i$ to denote scalars. Operators $(\cdot)^\text{T}$, $\left\| \cdot \right\|_2$, $\text{E}\{\cdot\}$, $\text{col}\{\cdot\cdot\cdot\}$, $\text{diag}\{\cdot\cdot\cdot\}$, $\text{Tr}(\cdot)$, $\lambda_{\max} (\cdot)$, and $\otimes$ represent the transpose, $l_2$-norm of a vector, mathematical expectation, deployment of a vector by successively staking its arguments, diagonal or block diagonal operation over its arguments, trace of a matrix, maximum eigenvalue of a matrix, and Kronecker product of two matrices, respectively. $\text{vec}(\cdot)$ stacks the columns of an $L \times L$ matrix to form an $L^2 \times 1$ vector, and $\text{vec}^{-1}(\cdot)$ is its inverse operator. Also, $\bm I_L$ is an $L \times L$ identity matrix, and $\bm 1_N$ is a $N\times1$ vector with 1's value.
\section{Signal Model and \text{DNLMS} Algorithm}
Consider a connected network consisting of~$N$ sensor nodes geographically distributed. Every node $k$ communicates only with its single-hop neighbors and the communication between interconnected nodes is bidirectional. The set of single-hop neighbors to node $k$ (including itself) is denoted by~$\mathcal{N}_k$. At every time instant $i$, every node $k$ acquires a desired output scalar $d_k(i)$ and an $L\times1$ input regressor~$\bm u_{k,i}$, in which $d_k(i)$ and $\bm u_{k,i}$ are related by the linear model:
\begin{equation}
d_k(i) = \bm u_{k,i}^\text{T}\bm w^o + v_k(i),
\label{001}
\end{equation}
where $\bm w^o$ is an $L\times1$ sparse parameter vector and $v_k(i)$ is the additive noise at node \emph{k} independent of $\bm u_{m,j}$ for any $m$ and~$j$. The model~(\ref{001}) can be found in many applications~\cite{sayed2014adaptation, sayed2011adaptive}.

The objective of the network nodes is to use the streaming data $\{d_k(i),\bm u_{k,i}\}_{k=1}^N$ to perform the estimation of the vector $\bm w^o$ in a recursive way. To this end, the commonly used global MSE minimization problem\footnote{Another popular criteria is the exponentially weighted least squares in the network: $\min \limits_{\bm w} \left\lbrace  \lambda^{i+1}\delta\lVert \bm w\rVert_2^2 + \sum\limits_{j=0}^i \lambda^{i-j} \sum\limits_{k=1}^N \left(d_k(j)-\bm u_{k,j}^T\bm w\right)^2 \right\rbrace$~\cite{chu2017variable}.} is stated as
\begin{equation}
\label{001afA}
\begin{array}{rcl}
\begin{aligned}
\min \limits_{\bm w}  \sum_{k=1}^N \text{E} \left\lbrace (d_k(i) - \bm u_{k,i}^\text{T}\bm w)^2\right\rbrace.
\end{aligned}
\end{array}
\end{equation}
Based on the diffusion strategy that every node fuses linearly its own information and the received information from its neighbors,~\eqref{001afA} is equivalent to minimizing the local MSE cost functions for all the nodes $k=1,...,N$~\cite{sayed2014adaptation,cattivelli2010diffusion}:
\begin{equation}
\label{001afB}
\begin{array}{rcl}
\begin{aligned}
&\min \limits_{\bm w_k} J_k^{loc}(i),\\
&J_k^{loc}(i) = \sum\limits_{m\in \mathcal{N}_k} c_{m,k} \text{E} \left\lbrace(d_m(i) - \bm u_{m,i}^\text{T}\bm w_k)^2\right\rbrace,
\end{aligned}
\end{array}
\end{equation}
where $c_{m,k}$ represents a weight that node $k$ assigns to the information coming from node~$m$, also called the combination coefficients. Note that $\{c_{m,k}\}$ requires $c_{m,k}\geq0$, $c_{m,k}=0$ if $m\notin\mathcal{N}_k$, and $\sum_{m\in\mathcal{N}_k}c_{m,k} =1$~\cite{takahashi2010diffusion}.

By using the SGD rule to solve~\eqref{001afB}, the DLMS algorithm~\cite{lopes2008diffusion} is obtained as
\begin{subequations} \label{002}
    \begin{align}
    \bm \psi_{k,i+1} &= \bm w_{k,i} + \mu_k \bm u_{k,i} e_k(i), \label{002a}\\
    \bm w_{k,i+1} &= \sum\limits_{m\in \mathcal{N}_k} c_{m,k} \bm \psi_{m,i+1}, \label{002b}
    \end{align}
\end{subequations}
where at node $k$,
\begin{equation}
\label{003}
\begin{array}{rcl}
\begin{aligned}
e_k(i) = d_k(i)-\bm u_{k,i}^\text{T} \bm w_{k,i}
\end{aligned}
\end{array}
\end{equation}
denotes the output error and $\mu_k>0$ is a constant step size. Specifically, in the adaptation step~\eqref{002a}, each node~$k$ updates from the current estimate $\bm w_{k,i}$ to the intermediate estimate $\bm \psi_{k,i+1}$. Then, in the combination step~\eqref{002b}, each node~$k$ fuses all the intermediate estimates of nodes $m\in \mathcal{N}_k$ to yield an innovative estimate $\bm w_{k,i+1}$. To make the step size range independent of the covariance matrices of input regressors, the DNLMS algorithm~\cite{jung2015variable} modifies~\eqref{002a} to
\begin{equation}
\label{004}
\begin{array}{rcl}
\begin{aligned}
\bm \psi_{k,i+1} = \bm w_{k,i} + \mu_k \frac{\bm u_{k,i} e_k(i)}{\left\| \bm u_{k,i}\right\|_2^2}.
\end{aligned}
\end{array}
\end{equation}
where $0<\mu_k<2$ is to guarantee the algorithm convergence.

For the scenario that $v_k(i)$ contains impulsive noise, the measurements of $d_k(i)$ have many outliers with large amplitudes. However, the MSE criterion can not distinguish these outliers. Moreover, the impulsive noise appears randomly with a small probability or appears with a short duration of times. In this case, the DLMS and DNLMS algorithms will experience poor convergence or even divergence.
\section{Proposed Diffusion M-estimate Algorithms}
In this section, we present the derivations of the D-NLMM and D-SNLMM algorithms.
\subsection{Derivation of D-NLMM}
To estimate $\bm w^o$ in impulsive noise, we define the robust minimization problem for nodes $k=1,...,N$:
\begin{equation}
\label{005}
\begin{array}{rcl}
\begin{aligned}
&\min \limits_{\bm w_k} J_k^{loc}(i),\\
&J_k^{loc}(i) = \sum\limits_{m\in \mathcal{N}_k} c_{m,k} g_m^{-1} \text{E} \left\lbrace \varphi (d_m(i)-\bm u_{m,i}^\text{T}\bm w_k) \right\rbrace,
\end{aligned}
\end{array}
\end{equation}
where $g_m>0$ is a free-specified parameter, and $\varphi(x)$ is an M-estimate function on variable $x$. To ensure that $J_k^{loc}(i)$ converges to the minimum, $\varphi(x)$ is a continuous even function with the property $\varphi(x_1)\geq \varphi(x_2)>\varphi(0)\geq 0$ for $|x_1|>|x_2|>0$ and is sub-differentiable at least~\cite{li1998linear}. Moreover, to prevent from outliers, there is a positive number~$\xi^*$, and after $|x|>\xi^*$, the score function $\varphi'(x) \triangleq \frac{\partial \varphi(x)}{\partial x}$ holds that $|\varphi'(x)| \leq |\varphi'(\xi^*)|$, e.g., the below MH function for such $\varphi(x)$. In other words, $\varphi(x)$ is convex but may not be strictly convex at the points~$|x|=\xi^*$. Additionally, $\varphi'(x)$ may also equal a positive number multiplied by $\text{sign}(x)$, where $\text{sign}(\cdot)$ is the signum operator\footnote{$\text{sign}(x)=$ 1, 0, and $-1$ for $x>0$, $x=0$, and $x<0$.}; as an example, the well-known sign strategy is $\varphi(x)=|x|$ so that $\varphi'(x)=\text{sign}(x)$, which leads to the DSE-LMS algorithm~\cite{ni2016diffusion}.

At time instant~$i$, the instantaneous sub-gradient of~\eqref{005} with respect to $\bm w_k$ is formulated as
\begin{equation}
\label{006}
\begin{array}{rcl}
\begin{aligned}
\bigtriangledown_{\bm w} J_k^{loc}(i) \backsimeq -\sum\limits_{m\in \mathcal{N}_k} c_{m,k} g_m^{-1} \bm u_{m,i} \varphi' (d_m(i)-\bm u_{m,i}^\text{T}\bm w_{k,i}). \\
\end{aligned}
\end{array}
\end{equation}

Based on the SGD rule, the update equation for estimating $\bm w^o$ is established:
\begin{equation}
\label{006afterA}
\begin{array}{rcl}
\begin{aligned}
\bm w_{k,i+1} &= \bm w_{k,i} + \\
&\mu_k \sum\limits_{m\in \mathcal{N}_k} c_{m,k} g_m^{-1} \bm u_{m,i} \varphi' (d_m(i)-\bm u_{m,i}^\text{T}\bm w_{k,i}). \\
\end{aligned}
\end{array}
\end{equation}

Following the diffusion cooperation~\cite{cattivelli2010diffusion,di2013distributed}, at iteration~$i$, the current estimate $\bm w_{k,i}$ and new estimate $\bm w_{k,i+1}$ are given by
\begin{equation}
\label{006afterB}
\begin{array}{rcl}
\bm w_{k,i} = \sum\limits_{m\in \mathcal{N}_k} c_{m,k} \bm \psi_{m,i}
\end{array}
\end{equation}
and \eqref{002b}, respectively. By plugging them into \eqref{006afterA}, we obtain
\begin{equation}
\label{006afterC}
\begin{array}{rcl}
\begin{aligned}
\bm \psi_{m,i+1} = \bm \psi_{m,i} + \mu_m g_m^{-1} \bm u_{m,i} \varphi' (d_m(i)-\bm u_{m,i}^\text{T}\bm w_{k,i}). \\
\end{aligned}
\end{array}
\end{equation}
In~\eqref{006afterC}, although $\bm w_{k,i}$ is unavailable for node $m$, we can approximate it with $\bm w_{m,i}$ because both are estimates of $\bm w^o$. Also, we replace $\bm \psi_{m,i}$ with $\bm w_{m,i}$, since the latter contains more information through~\eqref{006afterB}. Under these considerations, we arrive at the recursion for the D-NLMM algorithm\footnote{By exchanging the order of steps~\eqref{007a} and~\eqref{007b}, the CTA-based {D-NLMM} algorithm can be naturally obtained.}:

\begin{subequations}
    \label{007}
    \begin{align}
    \bm \psi_{k,i+1} &= \bm w_{k,i} + \mu_k \frac{\bm u_{k,i} \varphi' (e_k(i))}{\left\| \bm u_{k,i}\right\|_2^2}, \label{007a}\\
    \bm w_{k,i+1} &= \sum\limits_{m\in \mathcal{N}_k} c_{m,k} \bm \psi_{m,i+1}. \label{007b}
    \end{align}
\end{subequations}
where we also choose $g_k = \left\| \bm u_{k,i}\right\|_2^2$ to yield~\eqref{007a}.

In adaptive filters, several M-estimate strategies have been studied for~$\varphi(x)$ such as the Huber function~\cite{kumar2015robust,li2018diffusion}, the MH function~\cite{YZhou2011}, and the Hampel's three-part redescending function~\cite{YuZouLeast2000}, to develop robust adaptive algorithms in impulsive noise. Due to the MH's simplicity, we focus on it for presenting the D-NLMM algorithm in distributed estimation\footnote{Other robust strategies summarized in~\cite{mandanas2016robust} can also be extended in a way that they can become different distributed algorithms.}. Interestingly, the algorithm using the MH function has also a comparable performance to that using either the Huber function or the Hampel's three-part redescending function, as can be seen in Fig.~\ref{Fig3}.

The MH is a piecewise continuous function:
\begin{equation}
\label{008}
\varphi(e_k)=\left\{ \begin{aligned}
& e_k^2/2, \text{ if } |e_k| < \xi_k\\
&\xi_k^2/2, \text{ if } |e_k| \geq \xi_k,
\end{aligned} \right.
\end{equation}
and its score function is
\begin{equation}
\label{009}
\varphi'(e_k)=\left\{ \begin{aligned}
& e_k, \text{ if } |e_k| < \xi_k\\
&0,\;\; \text{ if } |e_k| \geq \xi_k,
\end{aligned} \right.
\end{equation}
where $\xi_k$ is a threshold. By combining~\eqref{007} and~\eqref{009}, it turns out that, at time instant $i$, when the magnitude of~$e_k(i)$ is smaller than $\xi_k$, $\varphi'(e_k(i))$ is equal to $e_k(i)$, and the proposed algorithm performs the DNLMS update. When $|e_k(i)| \geq \xi_k$ (which means the appearance of impulsive noise), $\varphi'(e_k(i))$ will become zero, thereby stopping the adaptation of the algorithm. Towards this goal, the threshold $\xi_k$ is adaptively adjusted by
\begin{equation}
\label{010}
\begin{array}{rcl}
\begin{aligned}
\xi_k =\kappa \sigma_{e,k}(i),
\end{aligned}
\end{array}
\end{equation}
where $\sigma_{e,k}^2(i)$ is the variance of $e_k(i)$ excluding impulsive noise. Typically, $\kappa=2.576$ for the suppression of impulsive noise, which means, under the assumption that $e_k(i)$ is Gaussian distributed except when being polluted accidentally by impulsive noise, the confidence level of preventing $e_k(i)$ from contributing to the update is 99\% when $|e_k| \geq \xi_k$~\cite{YuZouLeast2000}. And, $\sigma_{e,k}^2(i)$ can be estimated by the following recursion:
\begin{equation}
\label{011}
\begin{array}{rcl}
\begin{aligned}
\hat{\sigma}_{e,k}^2(i) = \zeta \hat{\sigma}_{e,k}^2(i-1) + (1-\zeta) \text{med}(\bm A_{k,i}^e),
\end{aligned}
\end{array}
\end{equation}
where $0 < \zeta \lesssim 1$ is a forgetting factor except $\zeta=0$ at the starting time $i=0$, $\text{med}(\cdot)$ is the median operator of the error data sliding window~$\bm A_{k,i}^e=[e_k^2(i),e_k^2(i-1),...,e_k^2(i-N_w+1)]$ which helps to avoid the effect of impulsive noise on~$\hat{\sigma}_{e,k}^2(i)$. The window length $N_w$ is usually chosen between 5 and~9; also, it should be increased appropriately when the occurrence probability of impulsive noise is high.

\subsection{Derivation of D-SNLMM}
In order to enforce the sparsity of $\bm w^o$, we propose to incorporate a real-valued sparse regularization $F(\bm w_k)$ on $\bm w_k$ into~\eqref{005} and obtain:
\begin{equation}
\label{012}
\begin{array}{rcl}
\begin{aligned}
&\min \limits_{\bm w_k} J_k^{loc}(i),\\
&J_k^{loc}(i) = \sum\limits_{m\in \mathcal{N}_k} c_{m,k} g_m^{-1} \text{E} \left\lbrace \varphi (d_m(i)-\bm u_{m,i}^\text{T}\bm w_k) \right\rbrace +\beta F(\bm w_k) \\
\end{aligned}
\end{array}
\end{equation}
for nodes $k=1,2,...,N$, where the regularization parameter~$\beta>0$ controls the intensity given to $F(\bm w_k)$.

By following the derivation procedure in the above subsection to~\eqref{012}, the D-SNLMM algorithm for estimating $\bm w^o$ is formulated as
\begin{subequations} \label{013}
    \begin{align}
    \bm \psi_{k,i+1} &= \bm w_{k,i} + \mu_k \frac{\bm u_{k,i} \varphi' (e_k(i))}{\left\| \bm u_{k,i}\right\|_2^2} - \mu_k \beta f(\bm w_{k,i}), \label{013a}\\
    \bm w_{k,i+1} &= \sum\limits_{m\in \mathcal{N}_k} c_{m,k} \bm \psi_{m,i+1}, \label{013b}
    \end{align}
\end{subequations}
where $f(\bm w_{k,i})$ is referred to as the zero attractor:
\begin{equation}
\label{014}
\begin{array}{rcl}
\begin{aligned}
f(\bm w_{k,i}) \triangleq \frac{\partial F(\bm w_{k,i})}{\partial \bm w_{k,i}} = \left[ f([\bm w_{k,i}]_1) ,..., f([\bm w_{k,i}]_L)  \right]^\text{T}
\end{aligned}
\end{array}
\end{equation}
where $[\bm w_{k,i}]_l$ is the $l$-th element of the vector $\bm w_{k,i}$.
\begin{table*}[tbp]
    \scriptsize
    \centering
    \vspace{-1em}
    \caption{Existing Functions for Approximating the $l_0$-norm.}
    \label{table_1}
    \begin{tabular}{@{}l|ccc}
        \hline
        \textbf{} &regularization function: $F(\bm w)$ &zero attractor: $f([\bm w]_l),\;l=1,...,L$\\
        \hline
        (a) \cite{liu2012diffusion,yu2018sparsity}& $\sum \limits_{l=1}^L |[\bm w]_l|$ &$\text{sign}([\bm w]_l)$\\
        (b)\cite{liu2012diffusion,yu2018sparsity}& $\sum \limits_{l=1}^L \frac{|[\bm w]_l|}{\varepsilon +|[\bm w]_l|}$ &$\frac{\text{sign}([\bm w]_l)}{\varepsilon +|[\bm w]_l|}$\\
        (c) \cite{gu2009l,olinto2016transient}& $\sum \limits_{l=1}^L \left( 1-\text{exp}^{-\upsilon|[\bm w]_l|}\right) $ &$\upsilon\text{sign}([\bm w]_l) \text{exp}^{-\upsilon|[\bm w]_l|}$ &\\
        (d) \cite{wu2013gradient}& $\sum \limits_{l=1}^L |[\bm w]_l|^p$ with a variable $p$ in $0<p<1$ & $\frac{p\text{sign}([\bm w]_l)}{\varepsilon +|[\bm w]_l|^{1-p}}$&\\
        (e) \cite{lima2013affine} &$\sum \limits_{l=1}^L \left( 1-\text{exp}^{-\frac{1}{2}\upsilon^2|[\bm w]_l|^2}\right) $ &$\upsilon^2 [\bm w]_l \text{exp}^{-\frac{1}{2}\upsilon^2|[\bm w]_l|^2}$\\
        (f) \cite{ShravanPoly2018} & $\sum \limits_{l=1}^L \left\{ \begin{aligned}
        &\frac{|[\bm w]_l|}{(1+|[\bm w]_l|)^\upsilon}, \text{ if } |[\bm w]_l| \leq \frac{1}{\upsilon-1}\\
        &\frac{|\upsilon-1|^{\upsilon-1}}{|\upsilon|^\upsilon}, \text{elsewhere},
        \end{aligned} \right.$ &$ \left\{ \begin{aligned}
        &\frac{\text{sign}([\bm w]_l)[1-(\upsilon-1)|[\bm w]_l|]}{(1+|[\bm w]_l|)^{\upsilon+1}}, \text{ if } |[\bm w]_l| \leq \frac{1}{\upsilon-1}\\
        &0, \text{elsewhere},
        \end{aligned} \right.$ \\
        \hline
    \end{tabular}
\end{table*}

For a sparse vector $\bm w^o$, only a few elements have large magnitudes while the remaining ones are zero. Intuitively, the $l_0$-norm should be a good metric for the sparsity of $\bm w^o$, namely, $||\bm w^o||_0$ which denotes the number of nonzero elements. Although we can not know the location of the non-zero elements beforehand, in some applications, we may have the priori knowledge on the upper bound of $||\bm w^o||_0$, i.e., $||\bm w^o||_0\leq s_{up}$. Unfortunately, since the $l_0$-norm is neither continuous nor differentiable, the $l_0$-norm based minimization is a NP-hard problem. As such, much literature has studied several functions approximating the $l_0$-norm as summarized in Table~\ref{table_1}, and developed different sparsity-aware algorithms. Among them, the approximation functions (a), (b), and (c) are probably the most widely used, due to their relative simplicity. The approximation function~(a) is also called the $l_1$-norm. As reported in the literature~\cite{liu2012diffusion,liu2014distributed,zheng2017diffusion}, using the function~(c) can make the algorithm better performance than using the function (a) or (b). Hence, we also consider the function~(c) in the D-SNLMM algorithm, i.e.,
\begin{equation}
\label{015}
\begin{array}{rcl}
\begin{aligned}
F(\bm w) = \sum \limits_{l=1}^L \left( 1-\text{exp}^{-\upsilon|[\bm w]_l|}\right).
\end{aligned}
\end{array}
\end{equation}
Note that, \eqref{015} is strictly equivalent to the $l_0$-norm when~$\upsilon \rightarrow \infty$. Thus, the zero attractor $f(\bm w)$ is given by
\begin{equation}
\label{016}
\begin{array}{rcl}
\begin{aligned}
f([\bm w]_l) = \upsilon\text{sign}([\bm w]_l) \text{exp}^{-\upsilon|[\bm w]_l|},\;l=1,...,L.
\end{aligned}
\end{array}
\end{equation}

Furthermore, by taking advantage of the first-order Taylor expansions of the exponential function, the low complexity version of~\eqref{016} is obtained:
\begin{equation}
\label{017}
f([\bm w]_l)=\left\{ \begin{aligned}
&- \upsilon^2 [\bm w]_l - \upsilon, \text{ if } -\frac{1}{\upsilon}\leq [\bm w]_l<0\\
&-\upsilon^2 [\bm w]_l + \upsilon, \text{ if } 0<[\bm w]_l \leq \frac{1}{\upsilon} \\
&0, \text{elsewhere}.
\end{aligned} \right.
\end{equation}

\emph{Remark 1:} The zero attractor $f(\bm w_{k,i})$ imposes an
attraction towards zero on small elements of the vector $\bm
w_{k,i}$ and those elements are in the majority, thereby bringing
about a performance improvement of the D-SNLMM algorithm when
estimating a sparse vector ${\bm w}^o$. As can be seen
from~\eqref{017} that the elements attracted are within a range of
$[a^1/\upsilon, 1/\upsilon]$, and the attraction intensity will be
greater if the element is closer to zero. It is worth noting that as
$\upsilon$ increases, the attraction intensity will become strong
but the attraction range will become narrow. Also, the proper choice
of $\beta$ will be explained later on in the analysis.
\begin{table*}[tbp]
    \scriptsize
    \centering
    \vspace{-1em}
    \caption{$\varphi(e_k(i))$ and $\varphi' (e_k(i))$ for Existing Robust Diffusion Algorithms.}
    \label{table_3}
    \begin{tabular}{@{}l|ccc}
        \hline
        \textbf{} &robust cost function: $\varphi(e_k(i))$ &the score function: $\varphi' (e_k(i))$\\
        \hline
        DSE-LMS~\cite{ni2016diffusion} & $|e_k(i)|$ &$\text{sign}(e_k(i))$\\
        DLMP~\cite{wen2013diffusion} & $|e_k(i)|^p$, where $1<p<2$ &$|e_k(i)|^{p-1}\text{sign}(e_k(i))$\\
        D-LLAD~\cite{chen2018diffusion}& $|e_k(i)|-\text{ln}(1+\alpha|e_k(i)|)$, where $\alpha>0$ &$\frac{\alpha|e_k(i)|}{1+\alpha|e_k(i)|}$ &\\
        DEN-LMS~\cite{al2017robust} &$\APLminus$ & $h_k(i)$ \\
        DNHuber~\cite{li2018diffusion} &$\left\{ \begin{aligned}
        &|e_k(i)|^2, \text{ if } |e_k(i)| < b\\
        &b|e_k(i)|-\frac{1}{2}b^2,\;\text{elsewhere}
        \end{aligned} \right. $ &$\left\{ \begin{aligned}
        &e_k(i), \text{ if } |e_k(i)| < b\\
        &b \text{sign}(e_k(i)),\;\text{elsewhere}
        \end{aligned} \right. $\\
        \hline
    \end{tabular}\\
    Note that: 1) $h_k(i)$ is a linear combination of preselected sign-preserving basis functions $\{\phi_{k,b}(e_k(i))$, $b=1,...,B_k\}$, where $B_k\geq1$.
    \\2) the parameters' notations (i.e., $p$, $\alpha$, $b$, and $B_k$) are the same as the ones in references.
\end{table*}

\begin{table}[tbp]
    \scriptsize
    \centering
    \vspace{-1em}
    \caption{Proposed D-SNLMM Algorithm and Its Special Versions.}
    \label{table_2}
    \begin{tabular}{lc}
        \hline
        \text{Initializations:} $\bm w_{k,0} = \bm 0$,  $\hat{\sigma}_{e,k}^2(0)=0$ \\
        \text{Parameters:} $\left. 0\ll \zeta<1\right. $, $0<\mu_k<2$, and $\beta \geq0 $\\
        \hline
        \text{D-SNLMM} algorithm: $g_{k,i} = \left\| \bm u_{k,i}\right\|_2^2$, $\beta > 0$\\
        \text{D-SLMM} algorithm: $g_{k,i} = 1$, $\beta > 0$\\
        \text{D-NLMM} algorithm: $g_{k,i} = \left\| \bm u_{k,i}\right\|_2^2$, $\beta=0$\\
        \text{D-LMM} algorithm: $g_{k,i} = 1$, $\beta=0$\\
        \hline
        \text{for} \text{iteration} $i\geq 0$ \text{do}\\
        \text{for} \text {each node \emph{k}} \text{do}\\
        \text{ }\text{ }\text{ }\text{ } $e_k(i) = d_k(i)-\bm u_{k,i}^\text{T} \bm w_{k,i}$\\
        \text{ }\text{ }\text{ }\text{ } $\left. \bm A_{k,i}^e =[e_k^2(i),e_k^2(i-1),...,e_k^2(i-N_w+1)]\right. $\\
        \text{ }\text{ }\text{ }\text{ } $\hat{\sigma}_{e,k}^2(i) = \zeta \hat{\sigma}_{e,k}^2(i-1) + (1-\zeta) \text{med}(\bm A_{k,i}^e)$ \\
        \text{ }\text{ }\text{ }\text{ } $\xi_k = 2.576\sigma_{e,k}(i)$\\
            \text{ }\text{ }\text{ }\text{ }$f([\bm w_{k,i}]_l)=\left\{ \begin{aligned}
        &- \upsilon^2 [\bm w_{k,i}]_l - \upsilon, \text{ if } -\frac{1}{\upsilon}\leq [\bm w_{k,i}]_l<0\\
        &-\upsilon^2 [\bm w_{k,i}]_l + \upsilon, \text{ if } 0<[\bm w_{k,i}]_l \leq \frac{1}{\upsilon} \\
        &0, \text{elsewhere}.
        \end{aligned} \right. $\\
        \text{ }\text{ }\text{ }\text{ }$f(\bm w_{k,i}) \triangleq \left[ f([\bm w_{k,i}]_1) ,..., f([\bm w_{k,i}]_L)  \right]^\text{T}$\\
        \text{ }\text{ }\text{ }\text{ } \text{if} $|e_k| < \xi_k$\\
        \text{ }\text{ }\text{ }\text{ } \text{} $\bm \psi_{k,i+1} = \bm w_{k,i} + \mu_k g_{k,i}^{-1} \bm u_{k,i} e_k(i) - \mu_k \beta f(\bm w_{k,i})$ \\
        \text{ }\text{ }\text{ }\text{ } \text{else} \\
        \text{ }\text{ }\text{ }\text{ } \text{} $\bm \psi_{k,i+1} = \bm w_{k,i} - \mu_k \beta f(\bm w_{k,i})$\\
        \text{ }\text{ }\text{ }\text{ } \text{end}\\
        \text{ }\text{end}\\
        \text{ }\text{for} \text {each node \emph{k}} \text{do}\\
        \text{ }\text{ }\text{ }\text{ } $\bm w_{k,i+1} = \sum\limits_{m\in \mathcal{N}_k} c_{m,k} \bm \psi_{m,i+1}$\\
        \text{ }\text{end}\\
        \text{end}\\
        \hline
    \end{tabular}
\end{table}

\emph{Remark 2}: The proposed D-NLMM and many existing robust
diffusion algorithms can be described in a unified
recursion~\eqref{007}, but the main difference is the score function
$\varphi'(e_k(i))$ to resist impulsive noise, which depends on the
specific robust strategy shown in Table~\ref{table_3}. In the
DEN-LMS algorithm, the score function is a weighted combination of
preselected sign-preserving basis functions and the weights need to
be optimized. In comparison, the D-NLMM algorithm is simpler in
implementation, since $\varphi'(e_k(i))$ only originates from the MH
function. Moreover, the D-NLMM algorithm considers the normalization
of input regressors in the adaptation. Based on the Huber function,
when $|e_k(i)| \geq b$, the DNHuber algorithm performs the
adaptation $\bm \psi_{k,i+1} = \bm w_{k,i} + \mu_k \frac{\bm u_{k,i}
\text{sign}((e_k(i))}{\left\| \bm u_{k,i}\right\|_2^2} $ rather than
the freezing like the D-NLMM algorithm. However, the D-NLMM
algorithm uses an adaptive threshold instead of the constant one in
the DNHuber algorithm so that the former outperforms the latter in
practice. Importantly, the D-SNLMM algorithm further improves the
D-NLMM performance in sparse parameter vector scenarios.
Particularly, if we set $g_k=1$, the non-normalized versions, i.e.,
the D-LMM and D-SLMM algorithms are obtained; they can be considered
as extensions of the MH function to the existing DLMS and ATC
$l_0$-LMS~\cite{liu2012diffusion} algorithms, respectively. The
proposed algorithms are summarized in Table~\ref{table_2}, where we
highlight the D-NLMM and D-SNLMM algorithms. We note that the
proposed algorithms can also be considered in the context of
detection problems
\cite{spa,mfsic,gmibd,tdr,mbdf,mmimo,lsmimo,armo,siprec,did,lclrbd,gbd,wlbd,bbprec,mbthp,rmbthp,badstbc,bfidd,1bitidd,baplnc,jpbnet}
and might be further enhanced by exploitation of low-rank techniques
\cite{intadap,inttvt,jio,ccmmwf,wlmwf,jidf,jidfecho,barc,jiols,jiomimo,jiostap,sjidf,l1stap,saabf,jioccm,ccmrab,wlbeam,lcrab,jiodoa,rrser,rcb,saalt,dce,damdc,locsme,memd,okspme,rrdoa,kaesprit,rhomo,sorsvd,corutv,sparsestap,wlccm,dcdrec,kacs}.

\section{Performance Analysis }
In this section, the mean and mean-square behaviors of the proposed algorithms in impulsive noise are studied. As stated in Remark~2, we will start the analysis from the D-SNLMM algorithm.

From \eqref{013}, we know that the difficulty of the performance analysis is how to evaluate the score term $\varphi'(e_k(i))$. Although the literature has studied the performance of the MH-based algorithms for adaptive filters~\cite{YZhou2011,Chan2010}, the extension to diffusion algorithms is not straightforward owing to the cooperation of interconnected nodes. More importantly, the existing analysis method for dealing with $\varphi'(e_k(i))$ is complicated, which involves the conditioned expectation, Price's theorem, and three complicated integrals. Therefore, we propose to move the score function $\varphi'(\cdot)$ out of~\eqref{013}, which makes the analysis simpler. Specifically, since the D-SNLMM algorithm performs the update of the estimate at node~$k$ when $|e_k(i)| < \xi_k$, we denote the probability of the update as
\begin{equation}
\begin{array}{rcl}
\begin{aligned}
\label{018}
P_{u,k}(i) = P\{|e_k(i)| < \xi_k\},
\end{aligned}
\end{array}
\end{equation}
whose calculation will be discussed in Section~IV.~C.
Accordingly, we can equivalently express~\eqref{013} in the mean as
\begin{subequations} \label{019}
\begin{align}
\bm \psi_{k,i+1} &= \bm w_{k,i} + \mu_k P_{u,k}(i) \frac{\bm u_{k,i} e_k(i) }{\left\| \bm u_{k,i}\right\|_2^2} - \mu_k \beta f(\bm w_{k,i}), \label{019a}\\
\bm w_{k,i+1} &= \sum\limits_{m\in \mathcal{N}_k} c_{m,k} \bm \psi_{m,i+1}. \label{019b}
\end{align}
\end{subequations}
It is noticed that \eqref{019} does not contain the score function $\varphi'(\cdot)$, which is absorbed into the evaluation of $P_{u,k}(i)$. This approach can also be extended to simplify the analyses of adaptive filtering algorithms in~\cite{YZhou2011,Chan2010}.

Subtracting $\bm w^o$ from both~\eqref{019a} and \eqref{019b}, respectively, which yields
\begin{subequations} \label{020}
\begin{align}
\widetilde{\bm \psi}_{k,i+1} &= \widetilde{\bm w}_{k,i} - \mu_k P_{u,k}(i) \frac{\bm u_{k,i} e_k(i)}{||\bm u_{k,i}||_2^2} + \mu_k \beta f(\bm w_{k,i}), \label{020a}\\
\widetilde{\bm w}_{k,i+1} &= \sum\limits_{m\in \mathcal{N}_k} c_{m,k} \widetilde{\bm \psi}_{m,i+1}, \label{020b}
\end{align}
\end{subequations}
where $\widetilde{\bm w}_{k,i} \triangleq \bm w^o - \bm w_{k,i}$ and $\widetilde{\bm \psi}_{k,i+1} \triangleq \bm w^o - \bm \psi_{k,i+1}$ indicate the error vector and the intermediate error vector, respectively. The relation \eqref{020} will be the starting point of performance analysis. For convenience of analysis, we make the following assumptions.

\emph{Assumption 1:} The regressors $\bm u_{k,i}$ are zero-mean with correlation matrices $\bm R_k=E\{\bm u_{k,i} \bm u_{k,i}^\text{T}\}$ in spatial independence.

\emph{Assumption 2:} The regressors $\bm u_{k,i}$ are independent of the estimation deviation $\widetilde{\bm w}_{m,i}$ for $j\leq i$ and all $k,m$. This is the well-known \emph{independence assumption} in the performance analyses of adaptive filtering algorithms \cite{sayed2011adaptive,7857046,chen2016generalized} and distributed algorithms~\cite{sayed2014adaptation}.

\emph{Assumption 3:} At every node $k$, the additive noise $v_k(i)$ includes the background noise $\theta_k(i)$ and the impulsive noise $\eta_k(i)$, namely, $v_k(i)=\theta_k(i) + \eta_k(i)$. The background noise $\theta_k(i)$ is drawn from a zero-mean white Gaussian process with variance~$\sigma_{\theta,k}^2$.

\emph{Assumption 4:} The impulsive noise $\eta_k(i)$ is modelled by the Bernoulli-Gaussian (BG) process: $\eta_k(i)=b_k(i) \cdot g_k(i)$, where $b_k(i)$ is a Bernoulli process whose pdf is expressed as $P[b_k(i)=1]=p_{k}$  and $P[b_k(i)=0]=1-p_{k}$, and $g_k(i)$ is drawn from a zero-mean white Gaussian process with variance $\sigma_{g,k}^2$, with $\sigma_{g,k}^2 \gg \sigma_{\theta,k}^2$. Note that, $p_{k}$ also stands for the probability of occurrence of impulsive noise.

According to assumptions 3 and 4, it is seen that the additive noise $v_k(i)$ is a CG process\footnote{In practice, the Alpha ($\alpha$)-stable process may be more efficient than the CG process for describing impulsive noise~\cite{georgiou1999alpha,LDang2019Kernel}, but it has no explicit expression for the pdf so that it is very difficult for performance analysis.} with zero-mean and variance $\sigma_{v,k}^2 = p_{k}\sigma_{s,k}^2 + (1-p_{k})\sigma_{\theta,k}^2$, where $\sigma_{s,k}^2 = \sigma_{g,k}^2 + \sigma_{\theta,k}^2$. The CG model is used frequently for analyzing the algorithms in impulsive noise~\cite{YZhou2011,ni2016diffusion,Chan2010}.

Under the condition of $P_{u,k}(i)$, we can use assumption~3 to get the relation $e_{k}(i) = \bm u_{k,i}^\text{T} \widetilde{\bm w}_{k,i} + \theta_k(i)$. Hence,~\eqref{020a} becomes
\begin{equation}
\begin{array}{rcl}
\begin{aligned}
\label{021}
\widetilde{\bm \psi}_{k,i+1} =& ( \bm I_L- \mu_k P_{u,k}(i) \bm A_{k,i} ) \widetilde{\bm w}_{k,i} - \\
&\mu_k P_{u,k}(i) \bm b_{k,i} + \mu_k \beta f(\bm w_{k,i}),
\end{aligned}
\end{array}
\end{equation}
where $\bm A_{k,i} = \frac{\bm u_{k,i} \bm u_{k,i}^\text{T}}{||\bm u_{k,i}||_2^2}$ and $\bm b_{k,i} = \frac{\bm u_{k,i} \theta_k(i) }{||\bm u_{k,i}||_2^2}$.

Some global quantities on all the nodes are defined as follows:
\begin{equation}
\begin{array}{rcl}
\begin{aligned}
\label{022}
\widetilde{\bm w}_i & \triangleq \text{col}\{\widetilde{\bm w}_{1,i}, ..., \widetilde{\bm w}_{N,i}\} \\
\widetilde{\bm \psi}_i & \triangleq\text{col}\{\widetilde{\bm \psi}_{1,i}, ..., \widetilde{\bm \psi}_{N,i}\} \\
\bm b_i & \triangleq\text{col}\{\bm b_{1,i},...,\bm b_{N,i}\} \\
\bm A_i & \triangleq\text{diag}\{\bm A_{1,i},...,\bm A_{N,i}\} \\
\mathcal{\bm M} & \triangleq\text{diag}\{\mu_1 \bm I_L,...,\mu_N \bm I_L\} \\
\mathcal{\bm P}_i & \triangleq\text{diag}\{P_{u,1}(i) \bm I_L,..., P_{u,N}(i)\bm I_L\} \\
f(\bm w_{i}) &\triangleq\text{col}\{f(\bm w_{1,i}),...,f(\bm w_{N,i})\}\\
\mathcal{\bm C} & \triangleq \bm C\otimes \bm I_L,\\
\end{aligned}
\end{array}
\end{equation}
where the matrix $\bm C$ collects all the combination coefficients~$\{c_{m,k}\}$, thus each column of $\bm C$ sums up to one (i.e., $\bm C^T \bm 1_N =\bm 1_N$). Using the above quantities, we rearrange~\eqref{021} and \eqref{020b} in a compact form:
\begin{equation}
\begin{array}{rcl}
\begin{aligned}
\label{023}
\widetilde{\bm w}_{i+1} =& \mathcal{\bm C}^\text{T} \widetilde{\bm \psi}_{i+1} \\
=& \mathcal{\bm C}^\text{T} (\bm I_{NL}- \mathcal{\bm M} \mathcal{\bm P}_i \bm A_i) \widetilde{\bm w}_i - \mathcal{\bm C}^\text{T} \mathcal{\bm M} \mathcal{\bm P}_i \bm b_i + \\& \beta \mathcal{\bm C}^\text{T} \mathcal{\bm M} f(\bm w_{i})\\
\end{aligned}
\end{array}
\end{equation}
which shows how the network error vector evolves over time.

\subsection{Mean Behavior}
By taking the expectation of both sides of~\eqref{023} under assumptions 2 and 3, we obtain that the mean of $\bm w_{k,i}$ evolves according to the recursion on the time instant $i$:
\begin{equation}
\begin{array}{rcl}
\begin{aligned}
\label{029}
\text{E}\{ \widetilde{\bm w}_{i+1} \} = \bm \Gamma_i E\{\widetilde{\bm w}_i\} + \beta \mathcal{\bm C}^\text{T} \mathcal{\bm M} \text{E}\{f(\bm w_i)\}, \\
\end{aligned}
\end{array}
\end{equation}
where
\begin{equation}
\begin{array}{rcl}
\begin{aligned}
\label{030}
\bm \Gamma_i = \mathcal{\bm C}^\text{T} (\bm I_{NL}- \mathcal{\bm M} \mathcal{\bm P}_i E\{\bm A_i\}).
\end{aligned}
\end{array}
\end{equation}

From~\eqref{029}, we have the following statement.

\newtheorem{theorem}{Theorem}
\begin{theorem}
The D-SNLMM algorithm converges in the mean if the step sizes are chosen to satisfy
\begin{equation}
\begin{array}{rcl}
\begin{aligned}
\label{031}
0<\mu_k < \frac{2}{P_{u,k}(i) \lambda_{\max}(E\{\bm A_{k,i}\}) },\; k=1,...,N.
\end{aligned}
\end{array}
\end{equation}
Furthermore, in the steady-state, the estimates across all nodes for this algorithm are biased with respect to~$\bm w^o$, i.e.,
\begin{equation}
\begin{array}{rcl}
\begin{aligned}
\label{032}
\text{E} \{\bm w_{k,\infty}\} = \bm w^o - \underbrace{\beta (\bm I_{NL} - \bm \Gamma_i )^{-1} \mathcal{\bm C}^\text{T} \mathcal{\bm M} \text{E}\{f(\bm w_{\infty})\}}\limits_{\text{bias}}\\
\end{aligned}
\end{array}
\end{equation}
\end{theorem}
for $k=1,...,N$.
\begin{IEEEproof}
Repeatedly iterating~\eqref{029}, we have
\begin{equation}
\begin{array}{rcl}
\begin{aligned}
\label{033}
\text{E}\{ \widetilde{\bm w}_{i+1} \} =& \prod\limits_{j=0}^i \bm \Gamma_j E\{\widetilde{\bm w}_0\} + \bm y_i, \\
\end{aligned}
\end{array}
\end{equation}
where
\begin{equation}
\begin{array}{rcl}
\begin{aligned}
\label{034}
\bm y_i = \beta \sum_{j=0}^{i} \left( \prod\limits_{s=i-j+1}^i \bm \Gamma_s \right) \mathcal{\bm C}^\text{T} \mathcal{\bm M} \text{E}\{f(\bm w_{i-j} )\}. \\
\end{aligned}
\end{array}
\end{equation}
Let us introduce the block-maximum-norm of the $MN \times MN$ matrix $\bm \Psi$ with block entries of size $M \times M$ each, which is defined as~\cite{sayed2014adaptation}:
\begin{equation}
\begin{array}{rcl}
\begin{aligned}
\label{035}
\left\| \bm \Psi \right\|_{b,\infty} &\triangleq  \max_{\bm x\neq \bm 0} \frac{\left\| \bm \Psi \bm x \right\|_{b,\infty}}{\left\| \bm x \right\|_{b,\infty}},\\
\left\| \bm x \right\|_{b,\infty} &\triangleq \max_{1 \leqslant k \leqslant N} \left\| \bm x_k \right\|_2,
\end{aligned}
\end{array}
\end{equation}
where $\bm x=\text{col}\{\bm x_1, ..., \bm x_N \}$ is an $MN \times 1$ vector with block entries $\{\bm x_k\}$ of size $M \times 1$ each. Thus, $||\mathcal{\bm C}^T ||_{b,\infty} =1$ holds. Then, enforcing the block-maximum-norm on both sides of~\eqref{033} yields:
\begin{equation}
\begin{array}{rcl}
\begin{aligned}
\label{036}
||\text{E}\{ \widetilde{\bm w}_{i+1} \}||_{b,\infty} \leq \left\| \prod\limits_{j=0}^i \bm \Gamma_j E\{\widetilde{\bm w}_0\} \right\| _{b,\infty} + || \bm y_i ||_{b,\infty},
\end{aligned}
\end{array}
\end{equation}
where
\begin{equation}
\begin{array}{rcl}
\begin{aligned}
\label{037}
\left\| \prod\limits_{j=0}^i \bm \Gamma_j E\{\widetilde{\bm w}_0\} \right\| _{b,\infty} &\leq \prod\limits_{j=0}^i ||\bm \Gamma_j||_{b,\infty} \cdot ||E\{\widetilde{\bm w}_0\}||_{b,\infty}\\
 \leq (\max_{i} &||\bm \Gamma_i||_{b,\infty})^{i} \cdot ||E\{\widetilde{\bm w}_0\}||_{b,\infty},
\end{aligned}
\end{array}
\end{equation}
and
\begin{equation}
\begin{array}{rcl}
\begin{aligned}
\label{038}
&|| \bm y_i ||_{b,\infty} \leq \beta \sum_{j=0}^{i}  (\max_{i-j+1\leq s \leq i} ||\bm \Gamma_s||_{b,\infty})^{j} \times \\
&\;\;\;\;\;\;\;\;\;\;\;\;\;\;\;\;\;\;||\mathcal{\bm C}^T ||_{b,\infty} \cdot ||\mathcal{\bm M} \text{E}\{f(\bm w_{i-j} )\}||_{b,\infty} \\
&\leq \beta \sum_{j=0}^{i}  (\max_{i} ||\bm \Gamma_i||_{b,\infty})^{j} \cdot ||\mathcal{\bm M} \text{E}\{f(\bm w_{i-j} )\}||_{b,\infty}.
\end{aligned}
\end{array}
\end{equation}
If we can ensure $||\bm \Gamma_i||_{b,\infty}<1$ for any~$i$, it is easy to check that as $i \rightarrow \infty$,~\eqref{037} will approach to zero so that~\eqref{036} will converge to
\begin{equation}
\begin{array}{rcl}
\begin{aligned}
\label{039}
||\text{E}\{ \widetilde{\bm w}_{i+1} \}&||_{b,\infty} \leq || \bm y_i ||_{b,\infty} \\
& \stackrel{(38)}{\leq} \beta \frac{(\max \limits_{1\leq k \leq N} \mu_k) \cdot \max \limits_{i} ||\text{E}\{f(\bm w_{i} )\}||_{b,\infty}}{1-(\max \limits_{i} ||\bm \Gamma_i||_{b,\infty})},
\end{aligned}
\end{array}
\end{equation}
where the term $\max \limits_{i} ||\text{E}\{f(\bm w_{i} )\}||_{b,\infty}$ is finite, because $f(\bm w_{k,i})$ given by \eqref{017} has bounded elements. Equation~\eqref{039} means that the algorithm is mean stable under the condition that
\begin{equation}
\begin{array}{rcl}
\begin{aligned}
\label{040}
||\bm \Gamma_i||_{b,\infty} & = \left\| \mathcal{\bm C}^\text{T} (\bm I_{NL}- \mathcal{\bm M} \mathcal{\bm P}_i E\{\bm A_i\}) \right\|_{b,\infty}\\
&\leq \left\| \mathcal{\bm C}^T \right\|_{b,\infty} \cdot \left\| \bm I_{NL}- \mathcal{\bm M} \mathcal{\bm P}_i E\{\bm A_i\} \right\|_{b,\infty} \\
&= \left\| \bm I_{NL}- \mathcal{\bm M} \mathcal{\bm P}_i E\{\bm A_i\} \right\|_{b,\infty} \\
&\stackrel{(a)}{=} \max_{1 \leqslant k \leqslant N}  \left\| \bm I_L - \mu_k P_{u,k}(i) E\{\bm A_{k,i}\} \right\|_{2} \\
&\leq 1,
\end{aligned}
\end{array}
\end{equation}
where the rationale behind~$(a)$ is the block diagonal property of the matrices $\bm I_{LN}$, $\mathcal{\bm M}$, $\mathcal{\bm P}_i$, and $E\{\bm A_{i}\}$. It follows from~\eqref{040} that the step sizes are bounded by~\eqref{031}. Also, when $i \rightarrow \infty$, we have $\text{E}\{ \widetilde{\bm w}_{i+1}\} = \text{E}\{ \widetilde{\bm w}_{i}\}$ so that~\eqref{032} can be deduced from~\eqref{029}, thereby completing the proof of Theorem 1.
\end{IEEEproof}

From Theorem~1 and by setting~$\beta=0$, we know that the mean convergence condition of the D-NLMM algorithm is also shown in~\eqref{031}, and this algorithm is unbiased for estimating $\bm w^o$ across all nodes in impulsive noise, i.e.,  $\text{E} \{\bm w_{k,\infty}\} = \bm w^o,\; k=1,...,N$.
\subsection{Mean-Square Behavior}
Let us define the covariance matrix of the network error vector $\widetilde{\bm w}_{i}$ as
\begin{equation}
\begin{array}{rcl}
\label{041}
\begin{aligned}
\mathcal{\bm W}_i &\triangleq \text{E} \{ \widetilde{\bm w}_{i} \widetilde{\bm w}_{i}^\text{T} \}
\end{aligned}
\end{array}
\end{equation}
where the $k$-th $L \times L$ diagonal block, denoted as $\bm W_{k,i} \triangleq \text{E} \{ \widetilde{\bm w}_{k,i} \widetilde{\bm w}_{k,i}^\text{T} \}$, is the covariance matrix of the error vector $\widetilde{\bm w}_{k,i}$ at node $k$. Then, post-multiplying \eqref{023} by its transpose, and then taking the expectation on both sides of the equation under assumptions 2 and~3, we find the recursive relation:
\begin{equation}
\begin{array}{rcl}
\begin{aligned}
\label{042}
\mathcal{\bm W}_{i+1}& = \mathcal{\bm C}^\text{T} \text{E}\{(\bm I_{NL}- \mathcal{\bm M} \mathcal{\bm P}_i \bm A_i) \mathcal{\bm W}_i (\bm I_{NL}- \mathcal{\bm M} \mathcal{\bm P}_i \bm A_i)^\text{T} \} \mathcal{\bm C}  \\
&+\mathcal{\bm C}^\text{T} \mathcal{\bm M} \mathcal{\bm P}_i \mathcal{\bm B} \mathcal{\bm P}_i \mathcal{\bm M} \mathcal{\bm C}+\beta \bm \Gamma_i \text{E}\{\widetilde{\bm w}_i f^\text{T}(\bm w_{i})\} (\mathcal{\bm C}^\text{T} \mathcal{\bm M})^\text{T}\\
&+\beta \mathcal{\bm C}^\text{T} \mathcal{\bm M} \text{E}\{f(\bm w_{i}) \widetilde{\bm w}_i^\text{T}\} \bm \Gamma_i^\text{T}\\
&+ \beta^2 \mathcal{\bm C}^\text{T} \mathcal{\bm M} \text{E}\{f(\bm w_{i}) f^\text{T}(\bm w_{i})\} (\mathcal{\bm C}^\text{T} \mathcal{\bm M})^\text{T},\\
\end{aligned}
\end{array}
\end{equation}
where
\begin{equation}
\begin{array}{rcl}
\begin{aligned}
\label{043}
&\mathcal{\bm B} \triangleq \text{E}\{\bm b_i \bm b_i^\text{T}\}= \\
&\text{diag} \left\lbrace \sigma_{\theta,1}^2 \text{E}\left\lbrace \frac{\bm u_{1,i}\bm u_{1,i}^\text{T}}{||\bm u_{1,i}||_2^4}\right\rbrace ,..., \sigma_{\theta,N}^2 \text{E}\left\lbrace \frac{\bm u_{N,i}\bm u_{N,i}^\text{T}}{||\bm u_{N,i}||_2^4} \right\rbrace \right\rbrace.  \\
\end{aligned}
\end{array}
\end{equation}

Enforcing the vectorization operation on both sides of~\eqref{042} and applying the Kronecker product property $\text{vec}(\bm X \bm \varSigma \bm Y) = (\bm Y^\text{T} \otimes \bm X) \text{vec}(\bm \varSigma)$ for any matrices $\{\bm X, \bm \varSigma, \bm Y\}$ of compatible dimensions~\cite{sayed2011adaptive}, we can establish that
\begin{equation}
\begin{array}{rcl}
\begin{aligned}
\label{044}
\text{vec} &(\mathcal{\bm W}_{i+1}) = \mathcal{\bm F}_i \text{vec} (\mathcal{\bm W}_{i}) + \\
&\;\;\;\;\;\; (( \mathcal{\bm C}^\text{T}\mathcal{\bm M} \mathcal{\bm P}_i) \otimes (\mathcal{\bm C}^\text{T} \mathcal{\bm M} \mathcal{\bm P}_i)) \text{vec} (\mathcal{\bm B}) + \\
&\;\;\;\;\;\;\beta (\mathcal{\bm C}^\text{T} \mathcal{\bm M} \otimes \bm \Gamma_i) \text{vec}(\text{E} \{ \widetilde{\bm w}_{i} f^\text{T}(\bm w_i) \}  ) + \\
&\;\;\;\;\;\;\beta (\bm \Gamma_i \otimes \mathcal{\bm C}^\text{T} \mathcal{\bm M}) \text{vec}( \text{E} \{ \widetilde{\bm w}_{i} f^\text{T}(\bm w_i) \}^\text{T} ) + \\
&\;\;\;\;\;\;\beta^2 (\mathcal{\bm C}^\text{T} \mathcal{\bm M} \otimes \mathcal{\bm C}^\text{T} \mathcal{\bm M} ) \text{vec}( \text{E} \{ f(\bm w_i) f^\text{T}(\bm w_i)\} ),\\
\end{aligned}
\end{array}
\end{equation}
where
\begin{equation}
\begin{array}{rcl}
\begin{aligned}
\label{045}
\mathcal{\bm F}_i = &(\mathcal{\bm C}^\text{T} \otimes \mathcal{\bm C}^\text{T}) \left[  \bm I_{N^2L^2} - \bm I_{NL} \otimes (\mathcal{\bm M} \mathcal{\bm P}_i \text{E}\{\bm A_i\}) -\right.  \\
& (\mathcal{\bm M} \mathcal{\bm P}_i \text{E}\{\bm A_i\}) \otimes \bm I_{NL} + \\
&\left.  ((\mathcal{\bm M} \mathcal{\bm P}_i) \otimes (\mathcal{\bm M} \mathcal{\bm P}_i)) \text{E} \{\bm A_i \otimes \bm A_i \} \right] .
\end{aligned}
\end{array}
\end{equation}
The mean-square-deviation (MSD) at node $k$ is defined as $\text{MSD}_k(i) \triangleq \text{Tr} \{\bm W_{k,i}\}$, and the network MSD over all the nodes is defined as $\text{MSD}_\text{net}(i)=\frac{1}{N}\sum_{k=1}^N \text{MSD}_k(i) = \text{Tr} \{\mathcal{\bm W}_{i}\}/N$ \cite{sayed2014adaptation}. Naturally, by giving $\beta=0$ and performing the inverse operator $\text{vec}^{-1}(\cdot)$,~\eqref{044} will model the MSD evolution behavior of the D-NLMM algorithm\footnote{The excess MSE (EMSE) evolution behavior of the algorithm can also be described by~\eqref{044} according to the definition~$\text{EMSE}_k(i) \triangleq \text{Tr} \{\bm W_{k,i} \bm R_{k}\}$. }. However, when this model is used for describing the MSD evolution behavior of the D-SNLMM algorithm, it still requires knowing the moments $\text{E}\{ f^\text{T}(\bm w_i) \}$, $\text{E} \{ \bm w_{i} f^\text{T}(\bm w_i) \}$, and $\text{E} \{ f(\bm w_i) f^\text{T}(\bm w_i)\}$ beforehand. To calculate them, we resort to two assumptions:

\emph{Assumption 5:} All entries in $\widetilde{\bm w}_{i}$ are Gaussian. The assumption is widely used and can be verified by the central limit theorem~\cite{yu2018sparsity,gu2009l,olinto2016transient}. Interested readers for its studies can refer to~\cite{bershad1989probability,wagner2011probability}. For the $l$-th entry of $\widetilde{\bm w}_{i}$, its mean and variance can be calculated by $x_l(i)=[\text{E} \{\widetilde{\bm w}_{i}\}]_l$ and $\sigma_{x,l}^2(i) = [\mathcal{\bm W}_i]_{l,l} - [\text{E} \{\widetilde{\bm w}_{i}\}]_l^2$, respectively, where $[\cdot]_{l,l}$ denotes the $l$-th diagonal entry of a matrix. Thus, the mean and variance of $[\bm w_i]_l$ are obtained as $\bar{x}_l(i) = [\bm 1_N \otimes \bm w^o]_l - x_l(i)$ and $\sigma_{x,l}^2(i)$, respectively.

\emph{Assumption 6:} When $k\neq m$ and $l\neq j$, the approximations $\text{E} \{[\bm w_{k,i}]_l [f(\bm w_{m,i})]_j\}\approx \text{E} \{[\bm w_{k,i}]_l\} \text{E} \{[f(\bm w_{m,i})]_j\}$ and $\text{E} \{[f(\bm w_{k,i})]_l [f(\bm w_{m,i})]_j\}\approx \text{E} \{[f(\bm w_{k,i})]_l\} \text{E} \{[f(\bm w_{m,i})]_j\}$ are made  \cite{4518487,olinto2016transient,yu2018sparsity}. Although this is a strong \textquotedblleft separable assumption\textquotedblright, it leads to the simplification of the analysis.

Note that, the above two assumptions hold in simulations, see Figs.~12 and~13.

Rewriting $\text{E} \{ \widetilde{\bm w}_{i} f^\text{T}(\bm w_i) \}$ yields
\begin{equation}
\begin{array}{rcl}
\begin{aligned}
\label{046}
\text{E} \{ \widetilde{\bm w}_{i} f^\text{T}(\bm w_i) \} = \bm w^o\text{E} \{ f^\text{T}(\bm w_i) \} - \text{E} \{ \bm w_{i} f^\text{T}(\bm w_i) \}.
\end{aligned}
\end{array}
\end{equation}

Based on~\eqref{017} and assumption 5, $\text{E} \{[f(\bm w_i)]_l\}$ can be computed as
\begin{equation}
\begin{array}{rcl}
\begin{aligned}
\label{047}
\text{E}\{f(x)\} =& \frac{1}{\sqrt{2\pi} \sigma_{x}} \int _{-\infty}^\infty f(x) \text{exp} ({-\frac{(x-\bar{x})^2}{2\sigma_{x}^2}})dx \\
=&\frac{\upsilon^2 \sigma_{x}}{\sqrt{2\pi}} ( a_1 - a_2 ) -\frac{\upsilon^2 \bar{x}}{2} ( b_1 + b_2) \\
&+\frac{\upsilon}{2} ( b_1  - b_2 + 2b_3 ),
\end{aligned}
\end{array}
\end{equation}
where $\text{erf}(x) \triangleq \frac{2}{\sqrt{\pi}} \int_{0}^{x}\text{exp} {(-t^2)}dt$, $a_1=\text{exp} ({-\frac{(1/\upsilon - \bar{x})^2}{2\sigma_{x}^2}})$, $a_2 = \text{exp} ({-\frac{(1/\upsilon + \bar{x})^2}{2\sigma_{x}^2}})$, $b_1 = \text{erf} ( \frac{1/\upsilon - \bar{x}}{\sqrt{2}\sigma_{x}})$, $b_2 = \text{erf} (\frac{1/\upsilon + \bar{x} }{\sqrt{2}\sigma_{x}}) $, and $b_3 = \text{erf} ( \frac{\bar{x}}{\sqrt{2}\sigma_{x}} )$.

We define the block matrices~$\bm \Xi_i \triangleq \text{E} \{ \bm w_{i} f^\text{T}(\bm w_i) \}$ and $\bm \Pi_i \triangleq \text{E} \{ f(\bm w_i) f^\text{T}(\bm w_i)\}$ which have $N^2$ blocks with every block size of~$L \times L$, where $\bm \Xi_{k,m,i} \triangleq \text{E} \{ \bm w_{k,i} f^\text{T}(\bm w_{m,i}) \}$ and $\bm \Pi_{k,m,i} \triangleq \text{E} \{ f(\bm w_{k,i}) f^\text{T}(\bm w_{m,i})\}$ are the $(k,m)$-th block of the $\bm \Xi_i$ and $\bm \Pi_i$, respectively.

For the $k$-th diagonal block matrices $\bm \Xi_{k,k,i}$ and $\bm \Pi_{k,k,i}$ with $k=1,...,N$, their off-diagonal entries can be directly given using assumption~6; based on assumption~5, their the $l$-th diagonal entries are computed respectively as follows:
\begin{equation}
\begin{array}{rcl}
\begin{aligned}
\label{048}
\text{E} \{ [&\bm w_{k,i}]_l [f^\text{T}(\bm w_{k,i})]_l \} \triangleq \text{E}\{xf(x)\} =\\
& \frac{1}{\sqrt{2\pi} \sigma_{x}} \int _{-\infty}^\infty xf(x) \text{exp} ({-\frac{(x-\bar{x})^2}{2\sigma_{x}^2}})dx =\\
&\frac{\upsilon^2 \sigma_{x}^2}{\sqrt{\pi}} \left( \frac{1/\upsilon - \bar{x}}{\sqrt{2}\sigma_{x}} a_1 + \frac{1/\upsilon + \bar{x}}{\sqrt{2}\sigma_{x}} a_2 \right) +\\
& \frac{\sqrt{2} \upsilon^2 \sigma_{x} \bar{x}}{\sqrt{\pi}} (a_1 + a_2) - \frac{ \upsilon \sigma_{x}}{\sqrt{2\pi}} (a_1 + a_2 -2a_3) - \\
&\frac{ \upsilon^2 \sigma_{x}^2 + \upsilon^2\bar{x}^2}{2} (b_1 +b_2) + \frac{\upsilon \bar{x}}{2} ( b_1  - b_2 + 2b_3 ),
\end{aligned}
\end{array}
\end{equation}

\begin{equation}
\begin{array}{rcl}
\begin{aligned}
\label{049}
\text{E} \{ &[f(\bm w_{k,i})]_l [f^\text{T}(\bm w_{k,i})]_l \} \triangleq \text{E}\{f^2(x)\} = \\
&\frac{1}{\sqrt{2\pi} \sigma_{x}} \int _{-\infty}^\infty f^2(x) \text{exp} ({-\frac{(x-\bar{x})^2}{2\sigma_{x}^2}} )dx =\\
&-\frac{\upsilon^4 \sigma_{x}^2}{\sqrt{\pi}} \left( \frac{1/\upsilon - \bar{x}}{\sqrt{2}\sigma_{x}} a_1 + \frac{1/\upsilon + \bar{x}}{\sqrt{2}\sigma_{x}} a_2 \right) -\\
&\frac{\sqrt{2} \upsilon^4 \sigma_{x} \bar{x}}{\sqrt{\pi}} (a_1 - a_2) + \frac{ \sqrt{2} \upsilon^3 \sigma_{x}}{\sqrt{\pi}} (a_1 + a_2 -2a_3) - \\
&\frac{ \upsilon^4\sigma_{x}^2 + \upsilon^4\bar{x}^2 + \upsilon^2}{2} (b_1 +b_2) - \upsilon^3 \bar{x} ( b_1  - b_2 + 2b_3),
\end{aligned}
\end{array}
\end{equation}
where $a_3 = \text{exp}(-\frac{\bar{x}^2}{2\sigma_{x}^2} ) $.

For the $(k,m)$-th off-diagonal block matrices $\bm \Xi_{k,m,i}$ and $\bm \Pi_{k,m,i}$, where $k\neq m \in \{1,...,N\}$, we can also use assumption~6 to compute their off-diagonal entries. Nevertheless, the \textquotedblleft separable assumption\textquotedblright\; can not be used for computing the $l$-th diagonal entries $\text{E} \{ [\bm w_{k,i}]_l [f^\text{T}(\bm w_{m,i})]_l \} $ and $\text{E} \{ [f(\bm w_{k,i})]_l [f^\text{T}(\bm w_{m,i})]_l \}$\footnote{Similar problem appeared in the $l_1$-norm based sparse distributed algorithm~\cite{8468982}, but it directly applies \textquotedblleft separable assumption\textquotedblright.}, since $\bm w_{k,i} $ and $\bm w_{m,i}$ are of significant similarity for estimating $\bm w^o$. In view of the difficulty for obtaining the joint probability density function of $[\bm w_{k,i}]_l $ and $[\bm w_{m,i}]_l$, we propose the following symmetric approximations:
\begin{equation}
\begin{array}{rcl}
\begin{aligned}
\label{050}
\text{E} &\{ [\bm w_{k,i}]_l [f^\text{T}(\bm w_{m,i})]_l \} = \text{E} \{ [\bm w_{m,i}]_l [f^\text{T}(\bm w_{k,i})]_l \} \\
&\approx \frac{\text{E} \{ [\bm w_{k,i}]_l [f^\text{T}(\bm w_{k,i})]_l \} + \text{E} \{ [\bm w_{m,i}]_l [f^\text{T}(\bm w_{m,i})]_l \}}{2},
\end{aligned}
\end{array}
\end{equation}

\begin{equation}
\begin{array}{rcl}
\begin{aligned}
\label{051}
\text{E} &\{ [f(\bm w_{k,i})]_l [f^\text{T}(\bm w_{m,i})]_l \} = \text{E} \{ [f(\bm w_{m,i})]_l [f^\text{T}(\bm w_{k,i})]_l \} \\
&\approx \frac{\text{E} \{ [f(\bm w_{k,i})]_l [f^\text{T}(\bm w_{k,i})]_l \} + \text{E} \{ [f(\bm w_{m,i})]_l [f^\text{T}(\bm w_{m,i})]_l \}}{2}.
\end{aligned}
\end{array}
\end{equation}

\begin{theorem}
Both D-NLMM and D-SNLMM algorithms in impulsive noise are mean-square stable if the step sizes satisfy
\begin{equation}
\begin{array}{rcl}
\begin{aligned}
\label{052}
0<\mu_k< \frac{2}{P_{u,k}(i)},\;k=1,...,N.
\end{aligned}
\end{array}
\end{equation}
\end{theorem}
\begin{IEEEproof}
See Appendix~A.
\end{IEEEproof}

Choosing the step sizes based on Theorem~2 and assuming the existence of~$(\bm I_{L^2N^2} -\mathcal{\bm F}_\infty)^{-1}$, we can take the limits of both sides of~\eqref{044} at $i \rightarrow \infty$ to yield
\begin{equation}
\begin{array}{rcl}
\begin{aligned}
\label{053}
\text{vec} &(\mathcal{\bm W}_{\infty}) = (\bm I_{L^2N^2} -\mathcal{\bm F}_\infty)^{-1} \times \\
&\left[ ((\mathcal{\bm C}^\text{T}\mathcal{\bm M} \mathcal{\bm P}_\infty) \otimes (\mathcal{\bm C}^\text{T} \mathcal{\bm M} \mathcal{\bm P}_\infty)) \text{vec}(\mathcal{\bm B}) +\right. \\
& \beta (\mathcal{\bm C}^\text{T} \mathcal{\bm M} \otimes \bm \Gamma_\infty) \text{vec}(\text{E} \{ \widetilde{\bm w}_{\infty} f^\text{T}(\bm w_\infty) \}  ) + \\
&\beta (\bm \Gamma_\infty \otimes \mathcal{\bm C}^\text{T}\mathcal{\bm M}) \text{vec}( \text{E} \{ \widetilde{\bm w}_{\infty} f^\text{T}(\bm w_\infty) \}^\text{T} ) + \\
&\left.  \beta^2 (\mathcal{\bm C}^\text{T} \mathcal{\bm M} \otimes \mathcal{\bm C}^\text{T} \mathcal{\bm M}) \text{vec}( \text{E} \{ f(\bm w_\infty) f^\text{T}(\bm w_\infty)\} )\right],
\end{aligned}
\end{array}
\end{equation}
which represents the steady-state MSD of the D-SNLMM algorithm. In addition, applying the relation $\text{Tr}(\bm X \bm Y)= \text{vec}(\bm X^\text{T})^\text{T} \text{vec}(\bm Y)$, the steady-state network MSD of the algorithm is also expressed as
\begin{equation}
\begin{array}{rcl}
\begin{aligned}
\label{054}
\text{MSD}&_{\text{net}}(\infty) = \frac{1}{N} \text{vec}(\bm I_{LN})^\text{T} (\bm I_{L^2N^2} -\mathcal{\bm F}_\infty)^{-1} \times \\
&(( \mathcal{\bm C}^\text{T}\mathcal{\bm M} \mathcal{\bm P}_\infty) \otimes (\mathcal{\bm C}^\text{T} \mathcal{\bm M} \mathcal{\bm P}_\infty)) \text{vec} (\mathcal{\bm B}) + \Delta_\infty,
\end{aligned}
\end{array}
\end{equation}
where
\begin{equation}
\begin{array}{rcl}
\begin{aligned}
\label{055}
\Delta_\infty =
&\frac{1}{N} \text{vec}(\bm I_{LN})^\text{T} (\bm I_{L^2N^2} -\mathcal{\bm F})^{-1} \times  \\
&\left[  \beta (\mathcal{\bm C}^\text{T}\mathcal{\bm M} \otimes \bm \Gamma_\infty) \text{vec}(\text{E} \{ \widetilde{\bm w}_{\infty} f^\text{T}(\bm w_\infty) \}  ) + \right.\\
&\beta (\bm \Gamma_\infty \otimes \mathcal{\bm C}^\text{T} \mathcal{\bm M}) \text{vec}( \text{E} \{ \widetilde{\bm w}_{\infty} f^\text{T}(\bm w_\infty) \}^\text{T} ) + \\
&\left.  \beta^2 (\mathcal{\bm C}^\text{T} \mathcal{\bm M} \otimes \mathcal{\bm C}^\text{T} \mathcal{\bm M}) \text{vec}( \text{E} \{ f(\bm w_\infty) f^\text{T}(\bm w_\infty)\} )\right]
\end{aligned}
\end{array}
\end{equation}
is the result of the zero attractor characterizing the sparsity of~$\bm w^o$. Note that, if $\beta=0$ so that $\Delta_\infty=0$, \eqref{054} will reduce to the steady-state network MSD of the D-NLMM algorithm. Thus, by letting $\Delta_\infty<0$, we find the following theorem.
\begin{theorem}
For estimating sparse systems, the D-SNLMM algorithm has lower MSD than the D-NLMM algorithm only if the regularization parameter $\beta$ is in the range
\begin{equation}
\begin{array}{rcl}
\begin{aligned}
\label{056}
0<\beta <\beta^* \triangleq \frac{1}{N} \frac{\beta_a}{\beta_b},  \\
\end{aligned}
\end{array}
\end{equation}
where
\begin{equation}
\begin{array}{rcl}
\begin{aligned}
\label{057}
\beta_a = &-\text{vec}(\bm I_{LN})^\text{T} (\bm I_{L^2N^2} -\mathcal{\bm F})^{-1} \times \\
& \left[ (\mathcal{\bm C}^\text{T} \mathcal{\bm M}\otimes \bm \Gamma_\infty) \text{vec}(\text{E} \{ \widetilde{\bm w}_{\infty} f^\text{T}(\bm w_\infty) \}  ) + \right.\\
&\left. (\bm \Gamma_\infty \otimes \mathcal{\bm C}^\text{T}\mathcal{\bm M}) \text{vec}( \text{E} \{ \widetilde{\bm w}_{\infty} f^\text{T}(\bm w_\infty) \}^\text{T} ) \right],
\end{aligned}
\end{array}
\end{equation}
\begin{equation}
\begin{array}{rcl}
\begin{aligned}
\label{058}
\beta_b = &\text{vec}(\bm I_{LN})^\text{T} (\bm I_{L^2N^2} -\mathcal{\bm F})^{-1} \times \\
&(\mathcal{\bm C}^\text{T} \mathcal{\bm M} \otimes \mathcal{\bm C}^\text{T} \mathcal{\bm M} ) \text{vec}( \text{E} \{ f(\bm w_\infty) f^\text{T}(\bm w_\infty)\} ).
\end{aligned}
\end{array}
\end{equation}
\end{theorem}

\textit{Remark 3:} As shown in~\eqref{056}, the accuracy range of $\beta$ depends on the true~$\bm w^o$, but it can reveal the feasibility of the D-SNLMM algorithm outperforming the D-NLMM algorithm in sparse estimation problem. This phenomenon can also be observed in Figs.~\ref{Fig6} and~\ref{Fig7}. For the D-SNLMM algorithm,~\eqref{054} is an implicit equation on the steady-state MSD, but we may obtain its numerical solution by running the transient model~\eqref{044} to the steady-state.

\subsection{Calculation of $P_{u,k}(i)$}
To implement the above mean and mean-square models, now we will show how to compute the probability $P_{u,k}(i)$ defined in~\eqref{018}. Recalling the CG noise model and applying the law of total probability, we obtain
\begin{equation}
\begin{array}{rcl}
\begin{aligned}
\label{059}
P_{u,k}(i) =& p_{k} P\{|e_{s,k}(i)| < \xi_k\} + \\
&(1-p_{k}) P\{|e_{\theta,k}(i)| < \xi_k\},\\
\end{aligned}
\end{array}
\end{equation}
where $e_{s,k}(i) \triangleq \bm u_{k,i}^\text{T} \widetilde{\bm w}_{k,i} + \theta_k(i) + g_k(i)$ and $e_{\theta,k}(i) \triangleq \bm u_{k,i}^\text{T} \widetilde{\bm w}_{k,i} + \theta_k(i)$. It has been pointed out in \cite{al2003transient} that based on the central limit theorem, $\bm u_{k,i}^\text{T} \widetilde{\bm w}_{k,i}$ can be assumed to be Gaussian for $L\gg1$. Therefore, we are able to assume that $e_{s,k}(i)$ and $e_{\theta,k}(i) $ are zero mean Gaussian variables so that
\begin{equation}
\begin{array}{rcl}
\begin{aligned}
\label{060}
 P\{|e(i)| < \xi\} &\triangleq \frac{1}{\sqrt{2\pi}\sigma_{e}}\int_{-\xi}^{\xi} \text{exp} (-\frac{e^2}{2\sigma_{e}^2})de \\
&= \text{erf}(\xi/\sqrt{2}\sigma_{e}),\\
\end{aligned}
\end{array}
\end{equation}
where the subscripts $k$, $s$ and $\theta$ are omitted for concision. Subsequently,~\eqref{059} can be rewritten as
\begin{equation}
\begin{array}{rcl}
\begin{aligned}
\label{061}
P_{u,k}(i) =& p_{k} \text{erf}\left( \frac{\xi_k}{\sqrt{2}\sigma_{e_{s,k}}(i)}\right)  + (1-p_{k}) \text{erf}\left( \frac{\xi_k}{\sqrt{2}\sigma_{e_{\theta,k}}(i)}\right) ,\\
\end{aligned}
\end{array}
\end{equation}
where $\sigma_{e_{s,k}}^2(i)\triangleq \text{Tr}(\bm W_{k,i}\bm R_k) + \sigma_{s,k}^2$ and $\sigma_{e_{\theta,k}}^2(i)\triangleq \text{Tr}(\bm W_{k,i}\bm R_k) + \sigma_{\theta,k}^2$. Similar to \eqref{010}, $\xi_k$ is computed according to
\begin{equation}
\begin{array}{rcl}
\begin{aligned}
\label{062}
\xi_k =\kappa \sigma_{e_{\theta,k}}(i).\\
\end{aligned}
\end{array}
\end{equation}

In the steady-state, we introduce two approximations: $\sigma_{e_{\theta,k}}^2(\infty) \approx \sigma_{\theta,k}^2$ and $\sigma_{e_{s,k}}^2(\infty) \approx \sigma_{s,k}^2$, due to $\text{Tr} \left\lbrace \bm W_{k,i} \bm R_k \right\rbrace < \sigma_{\theta,k}^2 \ll \sigma_{s,k}^2$ for small step sizes. Consequently, $\mathcal{\bm P}_\infty$ in~\eqref{053} can be computed:
\begin{equation}
\begin{array}{rcl}
\begin{aligned}
\label{063}
P_{u,k}(\infty) =& p_{k} \text{erf}\left( \frac{ \kappa \sigma_{\theta,k} }{\sqrt{2} \sigma_{s,k} }\right)  + (1-p_{k}) \text{erf}\left( \frac{\kappa}{\sqrt{2} }\right).\\
\end{aligned}
\end{array}
\end{equation}

\subsection{Some Insights}
\textit{Remark 4:} From~\eqref{031} and~\eqref{052}, the step size range that guarantees the convergence of both the D-NLMM and D-SNLMM algorithms in both mean and mean-square senses is formulated as
\begin{equation}
\begin{array}{rcl}
\begin{aligned}
\label{064}
0<\mu_k &< \frac{1}{P_{u,k}(i)} \underbrace{\min \left\lbrace \frac{2}{\lambda_{\max}(E\{\bm A_{k,i}\})},\;2 \right\rbrace} \limits_{\text{DNLMS}} \\
&\stackrel{(b)}{\Rightarrow} 0<\mu_k < \frac{2}{P_{u,k}(i)}
\end{aligned}
\end{array}
\end{equation}
for all nodes $k=1,...,N$. It should be stressed that the term~$(b)$ stems from $\lambda_{\max}(E\{\bm A_{k,i}\}) \leq 1$. This is also because $E\{\bm A_{k,i}\}$ can also be considered as the normalized covariance matrix of input regressors, i.e., $E\{\bm A_{k,i}\}\approx \bm R_k/\text{Tr}(\bm R_k)$~when $L\gg1$~[53]. If $P_{u,k}(i)=1$, then~\eqref{064} reduces to the convergence condition of the DNLMS algorithm in the absence of impulsive noise. Because of $0<P_{u,k}(i)\lesssim 1$, one can infer from~\eqref{064} that the step size range of the D-NLMM algorithm is slightly wider than that of the DNLMS algorithm. More importantly, the convergence condition~\eqref{064} for the proposed algorithms is derived in the presence of impulsive noise.

\textit{Remark 5:} The previous analysis focuses on both the D-NLMM and D-SNLMM algorithms. However, following the previous analysis procedures, their non-normalized forms: the D-LMM and D-SLMM algorithms can also be easily analyzed, except the following differences:

1) In terms of the mean convergence condition, transient MSD, and steady-state MSD, we only need to replace $\text{E}\{\bm A_{i}\}$ with $\mathcal{\bm R}$, $\text{E} \{\bm A_i \otimes \bm A_i \}$ with $\mathcal{\bm R} \otimes \mathcal{\bm R}$, and $\mathcal{\bm B}$ with
\begin{equation}
\begin{array}{rcl}
\begin{aligned}
\label{065}
\mathcal{\bm B} = \text{diag} \left\lbrace \sigma_{\theta,1}^2 \bm R_1,..., \sigma_{\theta,N}^2 \bm R_N \right\rbrace, \\
\end{aligned}
\end{array}
\end{equation}
where $\mathcal{\bm R} = \text{diag} \{ \bm R_1,..., \bm R_N \}$.

2) On the mean-square convergence condition, $\bm \varSigma_k$ becomes
\begin{equation}
\begin{array}{rcl}
\begin{aligned}
\label{066}
\bm \varSigma_k \approx (\bm I_L- \mu_k P_{u,k}(i) \bm R_{k})^\text{T} (\bm I_L- \mu_k P_{u,k}(i) \bm R_{k}).
\end{aligned}
\end{array}
\end{equation}
From \eqref{066}, then we derive the bounds on step sizes that
\begin{equation}
\begin{array}{rcl}
\begin{aligned}
\label{067}
0<\mu_k < \frac{1}{P_{u,k}(i)} \cdot \underbrace{\frac{2}{\lambda_{\max}(\bm R_{k})} } \limits_{\text{DLMS}},\;k=1,...,N.
\end{aligned}
\end{array}
\end{equation}
Assuming $P_{u,k}(i)=1$,~\eqref{067} degrades into the step size range of the DLMS algorithm~\cite{tu2012diffusion}. From~\eqref{064} and~\eqref{067}, the convergence conditions of the D-NLMM and D-SNLMM algorithms do not depend on the maximum eigenvalues of the covariance matrices of input regressors, as opposed to the D-LMM and D-SLMM algorithms.

\section{Simulation Results}
\begin{figure}[htb]
\centering
\includegraphics[scale=0.55] {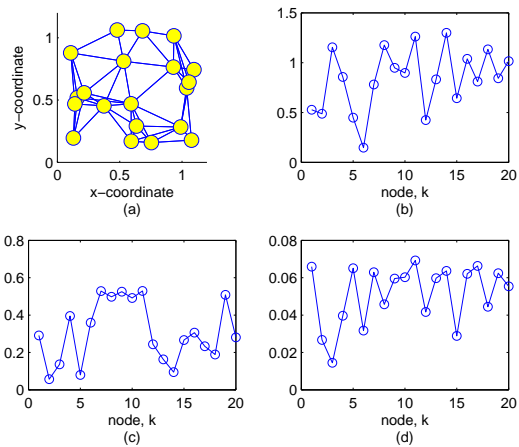}
\vspace{-1em} \caption{(a) Topology of network with 20 nodes, and values of (b) $\sigma_{\varepsilon,k}^2$ (c)~$\tau_k$ and (d) $\sigma_{\theta,k}^2$ at different nodes. }
\label{Fig1}
\end{figure}

Computer simulations are conducted over a distributed network with $N=20$ nodes (shown in Fig.~\ref{Fig1}(a), unless otherwise specified). The network MSD is used as a performance metric. All diffusion algorithms only consider the cooperation of the estimates in the combination step, and the combination coefficients $\{c_{m,k}\}$ are computed by the Metropolis rule~\cite{takahashi2010diffusion}:
\begin{equation*}
c_{m,k}= \left\{ \begin{aligned}
&1/\max(n_m,n_k),\;\text{if}\;m\in\mathcal{N}_k,\;m\neq k\\
&1-\sum \limits_{m\neq k}c_{m,k}, \text{if}\;m=k\\
&0,\;\text{otherwise},
\end{aligned} \right.
\end{equation*}
where $n_k$ is the number of neighbors of node $k$ including itself. All results are the average of 100 independent trials.
\subsection{Performance of Algorithms}
Let $Q$ denote the number of non-zero entries in the vector $\bm w^o$ of length $L=32$, and a smaller $Q$ means sparser $\bm w^o$. We set randomly values of non-zero entries from a Gaussian distribution, and then $\bm w^o$ is normalized by $\|\bm w^o\|_2=1$. The input regressor of node $k$ is given by $\bm u_{k,i}=[u_k(i),u_k(i-1),...,u_k(i-L+1)]^\text{T}$~\cite{LLi2010, ni2016diffusion}, with $u_k(i)$ being drawn from a first-order autoregressive model, $u_k(i)=\tau_k u_k(i-1) + \varepsilon_k(i)$, where $\varepsilon_k(i)$ is a zero mean white Gaussian process with variance $\sigma_{\varepsilon,k}^2$ and $\tau_k$ controls the correlation of $u_k(i)$ over time. Fig.~\ref{Fig1}(b) and (c) illustrate values of $\sigma_{\varepsilon,k}^2$ and $\tau_k$ at the nodes. The additive noise $v_k(i)$ interfering the desired output $d_k(i)$ at node~$k$ is drawn from a CG process described in assumptions~3 and~4, where Fig.~\ref{Fig1}(d) illustrates values of $\sigma_{\theta,k}^2$ for the Gaussian background noise.

\begin{figure}[htb]
    \centering
    \includegraphics[scale=0.52] {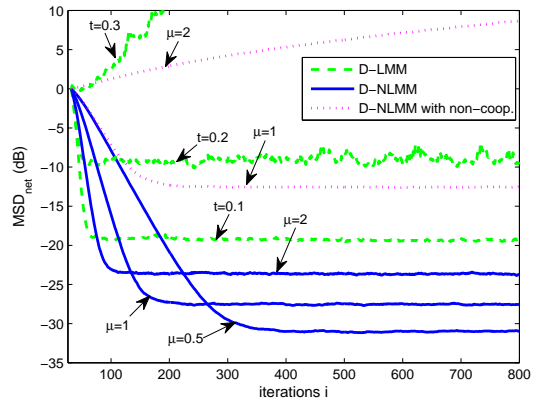}
    \vspace{-1em}\caption{Stability comparison of D-LMM and D-NLMM algorithms under CG noise with $p_k=0.01$ and $\sigma_{g,k}^2 = 10^4 \sigma_{\theta,k}^2$. Parameters setting of M-estimator is $N_w=9$, $\zeta=0.99$.}
    \label{Fig2}
\end{figure}

To begin with, we compare the stability of the D-LMM algorithm with that of the D-NLMM algorithm on step sizes. For conveniently calculating the maximum eigenvalue of the correlation matrix $\bm R_k$, we set Gaussian white input regressors for all~$k$, i.e., $u_k(i)=\varepsilon_k(i)$, which makes $\lambda_{\max}(\bm R_k)=\sigma_{\varepsilon,k}^2$. As shown in~\eqref{067}, the stability of the D-LMM algorithm is controlled by $\lambda_{\max}(\bm R_k)$ for all $k$. Thus, in this algorithm, we set the same step size at all the nodes to $\mu=t \cdot (2/\max \{ \sigma_{\varepsilon,k}^2|k=1,...,N\})$, where $0<t<1$. For the D-NLMM algorithm, we choose the same step size in the range~$0<\mu_k<2$ at all the nodes owing to $P_{u,k}(i)\lesssim 1$ in~\eqref{064}. The results for estimating non-sparse $\bm w^o$ are shown in Fig.~\ref{Fig2}. It is seen that unlike D-LMM, the stability range of D-NLMM on step sizes is not affected by the maximum eigenvalue of~$\bm R_k$ because it normalizes the adaptation process by the energies of input regressors. The D-NLMM with 'non-coop' is that every node runs independently an NLMM algorithm~\cite{Chan2010}, i.e., no cooperation among nodes. As we know, NLMM diverges when $\mu=2$. However, D-NLMM still converges when $\mu=2$, since~\eqref{052} is derived based on the relaxed inequalities~\eqref{0A7} and~\eqref{0A12}. Thanks to the cooperation of interconnected nodes, D-NLMM has much better performance than NLMM.

\begin{figure}[htb]
    \centering
    \includegraphics[scale=0.53] {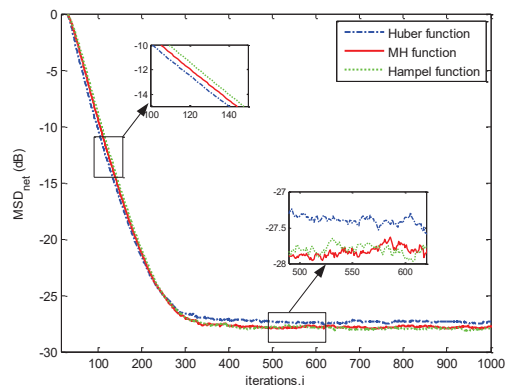}
    \vspace{-1em}\caption{Comparison of different M-estimate functions under CG noise with $p_k=0.01$ and $\sigma_{g,k}^2 = 10^4 \sigma_{\theta,k}^2$. [Sparse $\bm w^o$ with $Q=2$, $p_k=0.01$]. Parameters of M-estimate functions use their typical values in the literature. }
    \label{Fig3}
\end{figure}

Fig.~\ref{Fig3} shows the performance of the D-NLMM algorithm using different M-estimate functions. Unlike the Huber function used in the DNHuber algorithm, here all M-estimate functions are equipped with adaptive thresholds. It is clear that the MH and Hampel functions lead to similar but slightly superior performance to that of the Huber function. Among these M-estimators, the MH function is preferred due to its simplicity.

\begin{figure}[htb]
    \centering
    \includegraphics[scale=0.50] {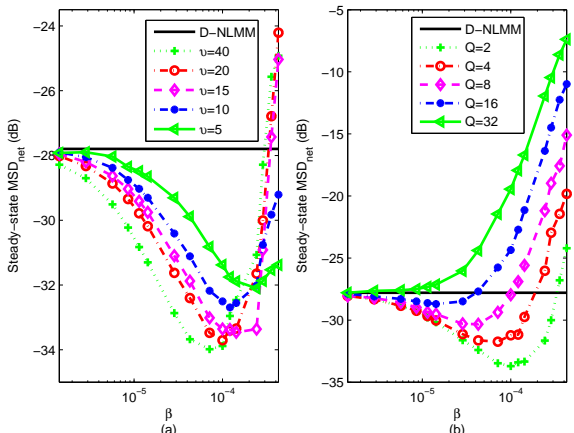}
    \vspace{-1em}
    \caption{Steady-state network MSDs of the D-SNLMM algorithm versus the parameter $\beta$, under CG noise with $p_k=0.01$ and $\sigma_{g,k}^2 = 10^4 \sigma_{\theta,k}^2$. (a) different values of $\upsilon$ and the fixed $Q=2$, (b) different sparsity degrees and the fixed $\upsilon=20$. }
    \label{Fig4}
\end{figure}

Then, Fig.~\ref{Fig4} examines the effect of the regularization parameter~$\beta$ on the steady-state performance of the D-SNLMM algorithm, where the D-NLMM algorithm ($\beta=0$) is used as a comparison benchmark when estimating the sparse vector~$\bm w^o$. The results are obtained by averaging over the last 100 MSD values after convergence to steady-state. As one can see, there is a region describing the choices of $\beta$ so that D-SNLMM has better steady-state performance than D-NLMM in the estimation of parameters with sparsity, as also indicated in~\eqref{056}. According to Fig.~\ref{Fig4}(a), a proper range of $\upsilon$ can be determined as~$\upsilon \in [15, 40]$. Moreover, in Fig.~\ref{Fig4}(b), as the sparsity degrees of $\bm w^o$ decrease (i.e., values of $Q$ increase), the superiority region on~$\beta$ for the D-SNLMM algorithm becomes narrow as compared to the D-NLMM algorithm, until it becomes null when $\bm w^o$ is non-sparse. That is to say, the D-SNLMM and D-NLMM algorithms are suitable for sparse and non-sparse scenarios, respectively.

\begin{figure}[htb]
    \centering
    \includegraphics[scale=0.53] {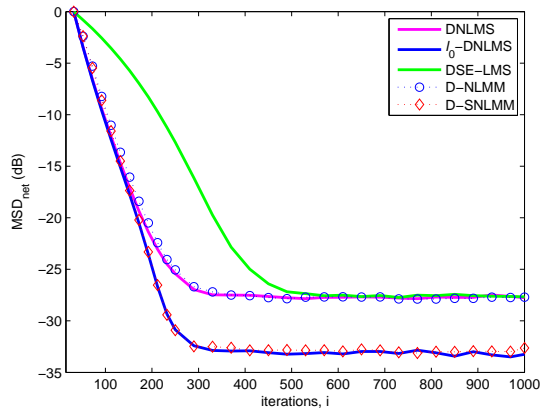}
    \vspace{-1em} \caption{Network MSD curves of diffusion algorithms under Gaussian noise. [Sparse $\bm w^o$ with $Q=2$]. Parameters of algorithms are set as follows: $\mu_k = 0.7$ (DNLMS); $\mu_k = 0.006$ (DSE-LMS); $\mu_k = 0.7$, $\alpha=20$, $\rho=6\times 10^{-5}$ ($l_0$-DNLMS); $\mu_k = 0.7$, $N_w=9$, $\zeta=0.99$ (D-NLMM), $\upsilon=20$, $\beta=8.6\times10^{-5}$ (D-SNLMM). }
    \label{Fig5}
\end{figure}

Fig.~\ref{Fig5} compares the performance of the DNLMS, $l_0$-DNLMS, DSE-LMS, and proposed D-NLMM and D-SNLMM algorithms in the case of no impulsive noise, where the $l_0$-DNLMS is the normalized version of the ATC $l_0$-LMS in~\cite{liu2012diffusion}. For a fair comparison, the parameters of algorithms are chosen based on the rule that the algorithms hold the same steady-state or convergence performance. As we can see from Fig.~\ref{Fig5}, the DSE-LMS algorithm is the slowest in the convergence. The D-NLMM and D-SNLMM algorithms keep almost the same performance as the DNLMS and $l_0$-DNLMS algorithms, respectively.

\begin{figure}[htb]
    \centering
    \includegraphics[scale=0.53] {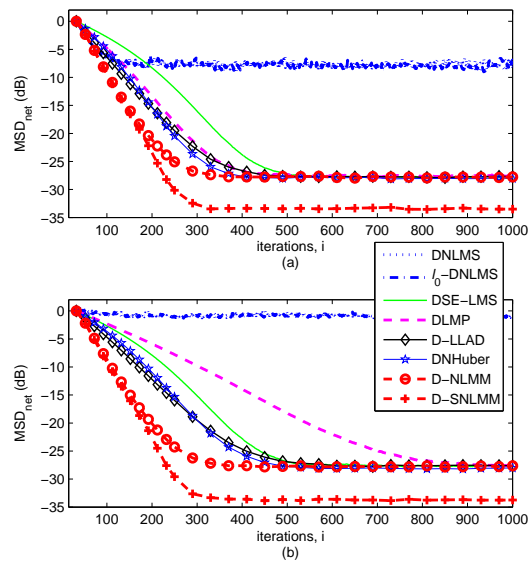}
    \vspace{-1em}
    \caption{Network MSD curves of diffusion algorithms under CG noise with $\sigma_{g,k}^2 = 10^4 \sigma_{\theta,k}^2$ and (a) $p_{k}=0.01$, (b) $p_{k}=0.05$. [Sparse $\bm w^o$ with $Q=2$]. Parameters of some algorithms are chosen as follows: $\mu_k = 0.0058$ (DSE-LMS); (a) $\mu_k = 0.01$ (b) 0.005, $p = 1.4$ (DLMP); (a) $\mu_k = 0.042$, $\alpha=0.5$ (b) $\mu_k = 0.018$, $\alpha=1.4$ (D-LLAD); $\mu_k = 0.7$, (a) $b = 0.4$ (b) $b = 0.3$ (DNHuber); (a) $N_w=9$ (b) $N_w=16$ (D-NLMM, D-SNLMM), and the remaining parameters are chosen as in Fig.~\ref{Fig5}.}
    \label{Fig6}
\end{figure}

In Fig.~\ref{Fig6}, we investigate the performance of the proposed algorithms in the presence of impulsive noise, where the occurrence probability of impulsive noise is set to $p_k=0.01$ and 0.05. The DLMP, D-LLAD, and DNHuber~\cite{li2018diffusion} algorithms are also required to a performance comparison. In this situation, the DNLMS and $l_0$-DNLMS algorithms have degraded performance, or even are of divergence in Fig.~\ref{Fig6}(b), yet other algorithms are insensitive to impulsive noise. For the DLMP algorithm, its performance is better when setting a smaller parameter $p$ in a strong impulsive noise case, as it becomes the DLMS and DSE-LMS algorithms when $p=2$ and $p=1$, respectively. The D-LLAD performs better than DSE-LMS but worse than DNHuber. Since D-NLMM employs the adaptive rule~\eqref{010} to select the threshold, it exhibits faster convergence rate than DNHuber using the fixed threshold. From Figs.~\ref{Fig5} and~\ref{Fig6}, one can also see that when identifying a sparse parameter vector~$\bm w^o$, the D-SNLMM reduces about 7~dB in the steady-state MSD as compared to that of D-NLMM, as the former takes advantage of the sparsity of $\bm w^o$.

\begin{figure}[htb]
    \centering
    \includegraphics[scale=0.53] {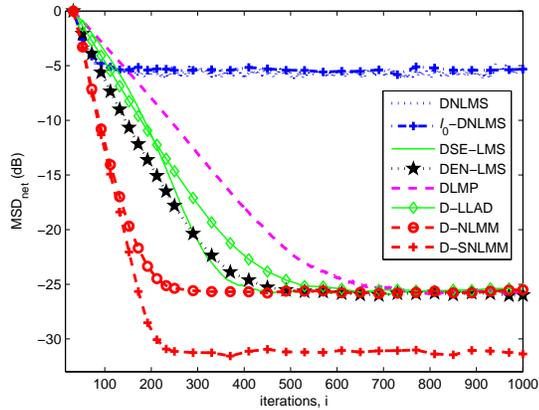}
    \vspace{-1em} \caption{Network MSD curves of diffusion algorithms under CG noise. [Sparse $\bm w^o$ with $Q=2$]. Parameters of algorithms are chosen as follows: $\mu_k = 0.008$ (DSE-LMS); $\mu_k = 0.007$, $p = 1.5$ (DLMP); $\mu_k = 0.024$, $\alpha=0.8$ (D-LLAD); $N_w=16$, $\zeta=0.95$ (M-estimator). The same step size $\mu_k=1$ is used for the DNLMS, $l_0$-DNLMS, D-NLMM, and D-SNLMM algorithms. The sparsity parameters of $l_0$-DNLMS and D-SNLMM are the same $\rho=\beta=1\times10^{-4}$.}
    \label{Fig7}
\end{figure}

In Fig.~\ref{Fig7}, we reset the CG noise parameters when the impulsive probability is increased to $p_k=0.1$ and the impulsive component $g_k(i)$ follows from the zero mean Laplacian distribution with variance $\sigma_{g,k}^2 = 10^3\sigma_{\theta,k}^2$. As in~\cite{al2017robust}, the DEN-LMS algorithm uses the basis functions~$\varphi_k^{tanh}=\{\phi_{k,1}=x, \phi_{k,2}=\tanh(x)\}$ and other parameters are chosen as $\mu_k=0.018$, $\nu_k=0.99$, and $\epsilon=10^{-6}$. The parameters of other algorithms are re-tuned to reach the same convergence or steady-state MSD as the DEN-LMS algorithm, except that both DNLMS and $l_0$-DNLMS algorithms have poor convergence in impulsive noise. As can be seen from Fig.~\ref{Fig7}, the convergence performance of DEN-LMS is superior to that of the DLMP and D-LLAD algorithms and approaches that of the DSE-LMS algorithm in the presence of Laplacian noise. However, the D-NLMM algorithm has faster convergence than than the DEN-LMS algorithm. Due to the use of sparsity, the D-SNLMM further improves the D-NLMM's steady-state performance.

\begin{figure}[htb]
    \centering
    \includegraphics[scale=0.53] {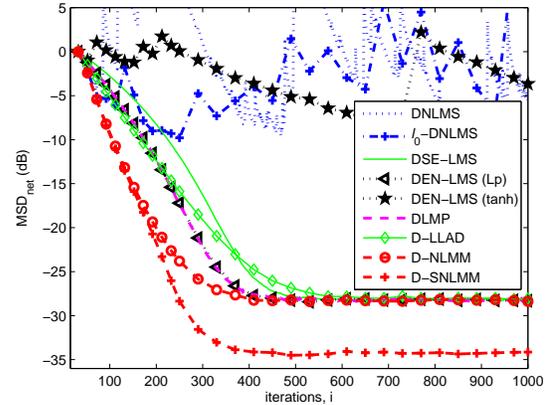}
    \vspace{-1em} \caption{Network MSD curves of diffusion algorithms in $\alpha$-stable noise. [Sparse $\bm w^o$ with $Q=2$]. Parameters of some algorithms are as follows: $\mu_k=0.006$ (DSE-LMS); $\mu_k=0.009$ (DLMP); $\mu_k = 0.03$, $\alpha=0.6$ (D-LLAD); $N_w=16$, $\zeta=0.95$ (M-estimator). The other parameters of the DNLMS, $l_0$-DNLMS, D-NLMM, and D-SNLMM algorithms are the same as Fig.~\ref{Fig5}.}
    \label{Fig8}
\end{figure}

On the other hand, the $\alpha$-stable process is used to model the additive noise $v_n$ with impulsive behavior, also called the $\alpha$-stable noise, whose characteristic function is expressed as~\cite{georgiou1999alpha}
\begin{align}
\Phi(t)=\exp(-\gamma \lvert t \lvert^\alpha).
\end{align}
The characteristic exponent $\alpha$ describes the impulsiveness of the noise (smaller $\alpha$ leads to more outliers) and $\gamma>0$ represents the dispersion degree of the noise. Specifically, when $\alpha$ = 1 or 2, it becomes the Cauchy noise or the Gaussian noise, respectively. In this example, we set $\alpha=1.3$ and $\gamma=2/15$. Fig.~\ref{Fig8} compares the performance of the previous algorithms. As we have known, in $\alpha$-stable noise environments, the least $p$-th ($Lp$) moment (where $p<\alpha$) is a proper criterion to devise the DLMP algorithm, and here we set $p=1.25$. Accordingly, we should reselect the basis functions for the DEN-LMS algorithm as $\varphi_k^{Lp}=\{\phi_{k,1}=x, \phi_{k,2}=|x|^{p-1}\text{sign}(x)\}$ and set the step size to $\mu_k=0.009$. It can be seen that the DEN-LMS version with the $\tanh$ basis exhibits poor convergence, like the DNLMS and $l_0$-DNLMS algorithms. By using the $Lp$ basis, the DEN-LMS version can converge as fast as the DLMP algorithm and outperforms the DSE-LMS and D-LLAD algorithms. However, among these robust algorithms, the proposed D-NLMM and D-SNLMM still achieve better estimation performance.

\subsection{Verification of Analysis}
In this subsection, the simulation setup is the same as in the above Fig.~\ref{Fig6}, unless otherwise specified. The distributed network has $N=10$ nodes and the length of $\bm w^o$ is $L=5$.

\begin{figure}[htb]
    \centering
    \includegraphics[scale=0.53] {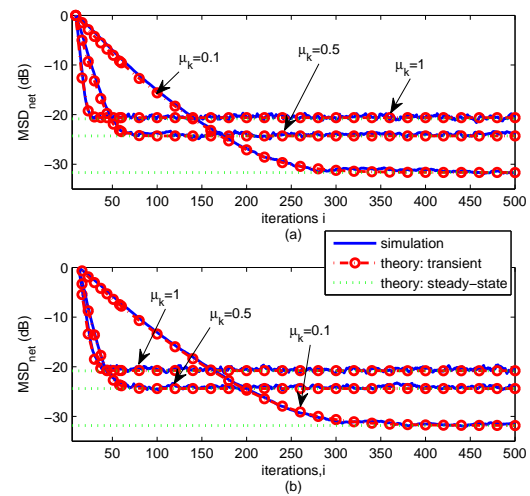}
    \vspace{-1em} \caption{Network MSD curves of the D-NLMM algorithm under CG noise with $\sigma_{g,k}^2 = 10^4 \sigma_{\theta,k}^2$. (a) $p_{k}=0.01$, (b) $p_{k}=0.05$.}
    \label{Fig9}
 \end{figure}

\begin{figure}[htb]
    \centering
    \includegraphics[scale=0.53] {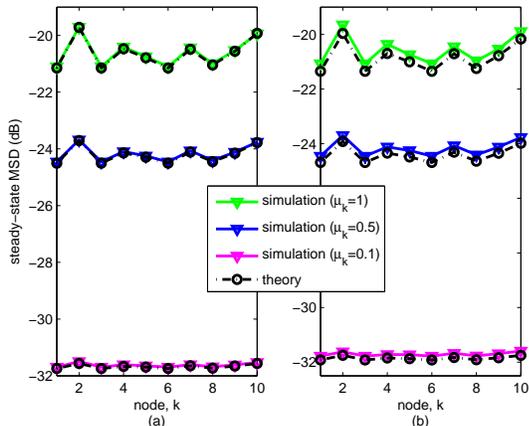}
    \vspace{-1em} \caption{Node-wise steady-state MSDs of the D-NLMM algorithm under CG noise with $\sigma_{g,k}^2 = 10^4 \sigma_{\theta,k}^2$. (a) $p_{k}=0.01$, (b) $p_{k}=0.05$.}
    \label{Fig10}
\end{figure}

In Figs.~\ref{Fig9} and~\ref{Fig10}, we check the analysis results of the D-NLMM algorithm for estimating the non-sparse vector $\bm w^o$ with $Q=5$. The theoretical transient results are computed by~\eqref{044}, and the theoretical steady-state results are computed by~\eqref{053} and~\eqref{063}, where $\beta=0$. Some expectations associated only with input regressors (e.g.,~$\bm R_k$) in the analyses are obtained by the ensemble average. Fig.~\ref{Fig9} shows the network MSD performance, and Fig.~\ref{Fig10} shows the steady-state MSDs at every node. As one can see, the theoretical results have a good agreement with the simulated results. For the node-wise steady-state results in Fig.~\ref{Fig10}(b), there is a small discrepancy when $\mu_k=0.5$ and 1, thanks mainly to the fact that~\eqref{063} is derived under small step sizes.

\begin{figure}[htb]
    \centering
    \includegraphics[scale=0.45] {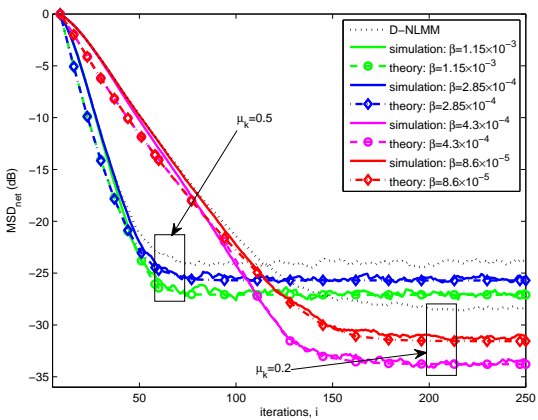}
    \vspace{-1em} \caption{Network MSD curves of the D-SNLMM algorithm under CG noise with $p_{k}=0.01$ and $\sigma_{g,k}^2 = 10^4 \sigma_{\theta,k}^2$.}
    \label{Fig11}
\end{figure}

\begin{figure}[htb]
    \centering
    \includegraphics[scale=0.5] {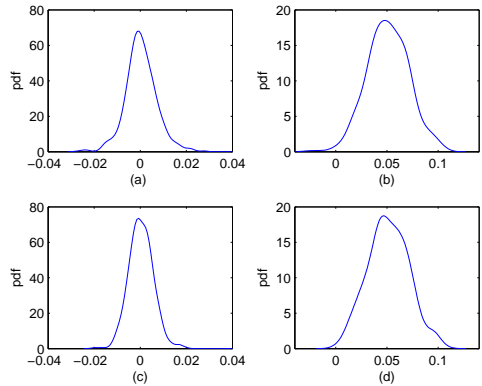}
    \vspace{-1em} \caption{Simulated pdf of the $l$-th entry in $\widetilde{\bm w}_{k,i}$ at iteration $i=100$. (a) $l=1$, $k=2$, (b) $l=4$, $k=2$, (c) $l=1$, $k=6$, (d) $l=4$, $k=6$. [The D-SNLMM algorithm chooses the parameters $\mu_k=0.2$, $\beta=3\times10^{-4}$].}
    \label{Fig12}
\end{figure}

\begin{figure}[htb]
    \centering
    \includegraphics[scale=0.5] {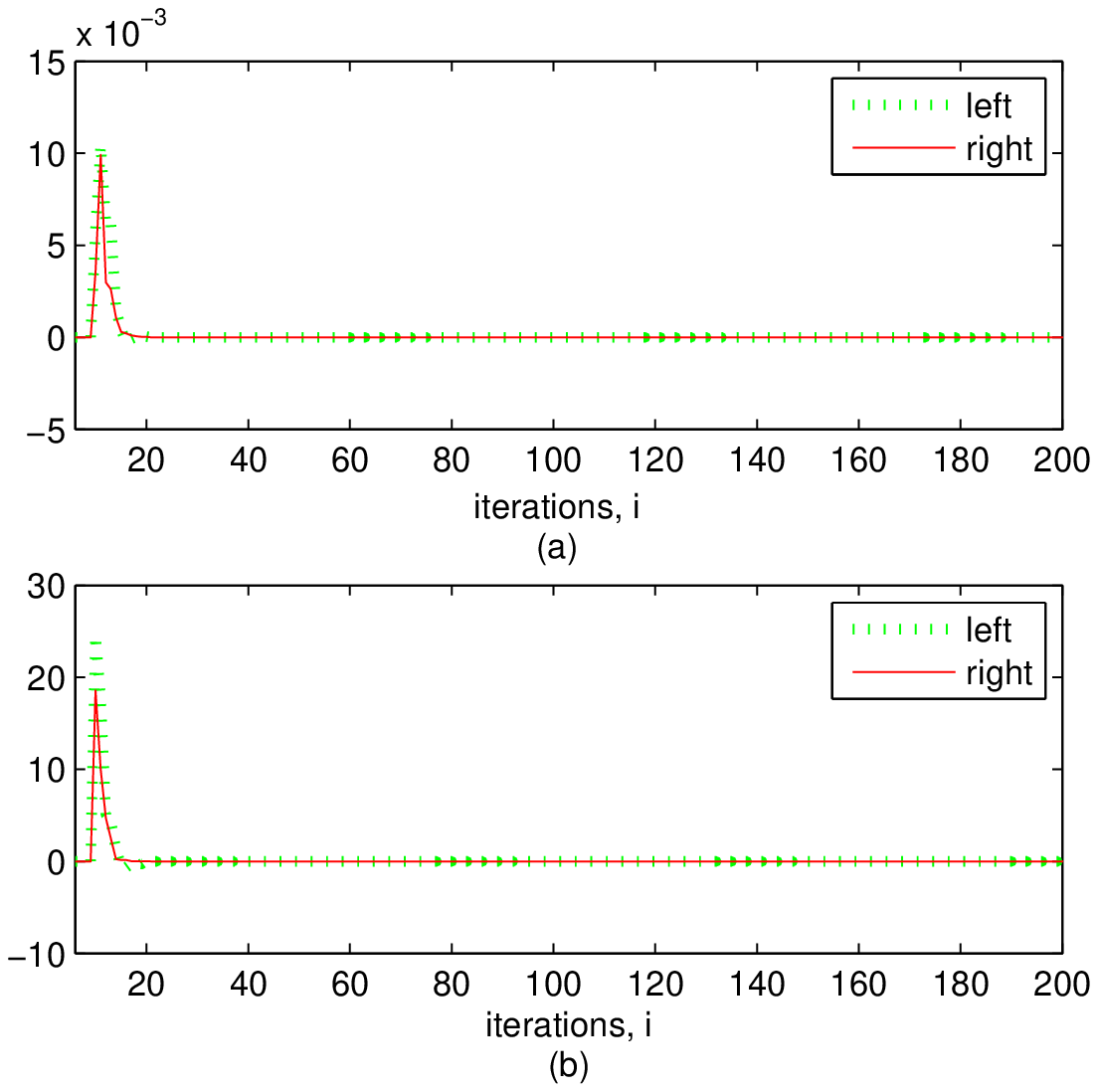}
    \vspace{-1em} \caption{Simulated results of assumption 6. (a) $\text{left}=\text{E} \{[\bm w_{2,i}]_1 [f(\bm w_{6,i})]_4\}$, $\text{right}=\text{E} \{[\bm w_{2,i}]_1\} \text{E} \{[f(\bm w_{6,i})]_4\}$, (b) $\text{left}=\text{E} \{[f(\bm w_{2,i})]_1 [f(\bm w_{6,i})]_4\}$, $\text{right}=\text{E} \{[f(\bm w_{2,i})]_1\} \text{E} \{[f(\bm w_{6,i})]_4\}$. }
    \label{Fig13}
\end{figure}

To test the theoretical model in~\eqref{044} for the D-SNLMM algorithm, the sparse vector is set as $\bm w^o=[0\;0\;0\;1\;0]^\text{T}$. The comparison results in impulsive noise are shown in Fig.~\ref{Fig11}. As can be seen, the difference between the theoretical results and the simulated results in the convergence stage of curves is relatively large, and it is negligible in the steady-state. The mechanism behind this difference is that assumptions~5 and~6 hold well in the steady-state. Furthermore, based on 500 independent trials, Fig.~\ref{Fig12} takes the simulated pdfs of the~$1$st and~$4$-th entries of the estimation error vector $\widetilde{\bm w}_{k,i}$ for this algorithm, at nodes $k=2,\;6$ and at iteration $i=100$ (transient). It is clear to see that for both zero and large coefficients in the sparse vector $\bm w^o$, the corresponding entries' estimation error approximately follow the Gaussian distribution. In the steady-state, the corresponding distribution is more like Gaussian, but here the figure is omitted. Fig.~\ref{Fig13} gives a simulated verification for assumption 6. From Figs.~\ref{Fig12} and~\ref{Fig13}, it is concluded that assumptions 5 and 6 can be applied to simplify the analysis of sparsity-aware algorithms.

\subsection{Comparison with Proximal Variants}
\begin{figure}[htb]
    \centering
    \includegraphics[scale=0.53] {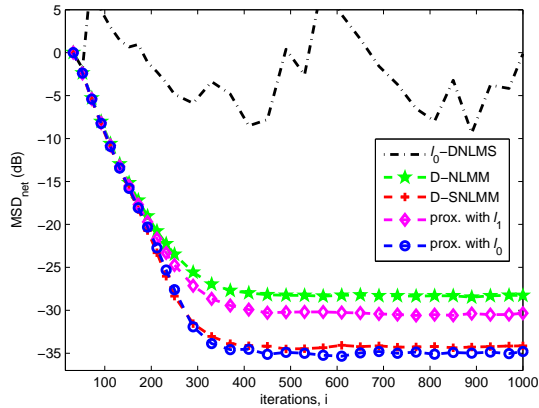}
    \vspace{-1em} \caption{Comparison of the D-SNLMM algorithm and the proximal variants under $\alpha$-stable noise. [Sparse $\bm w^o$ with $Q=2$].}
    \label{Fig14}
\end{figure}

Finally, Fig.~\ref{Fig14} compares the performance of the D-SNLMM algorithm with that of the proximal (prox.) variants with $l_1$-norm and $l_0$-norm (presented in Appendix~B) in $\alpha$-stable noise. These algorithms are employed to solve the distributed sparsity-aware minimization problem~\eqref{012}. The simulation setting is the same as Fig.~\ref{Fig8}, except $\beta=2\times10^{-4}$ for the 'prox. with $l_1$' variant and $\beta=6\times10^{-5}$ for the 'prox. with $l_0$' variant. As illustrated in Fig.~\ref{Fig14}, due to the $l_0$-norm based sparsity-regularization, both D-SNLMM and 'prox. with $l_0$' algorithms works better than the 'prox. with $l_1$' algorithm in terms of convergence and steady-state performance. Although the 'prox. with $l_0$' algorithm could slightly outperform the D-SNLMM algorithm, its complexity is slightly high. In particular, the additional complexity of the proximal algorithm originates from~\eqref{0B11}, i.e., $\text{prox}_{\mu_k\beta ||\cdot||_0}(\bm \Psi_{k,i+1}) = \max(|\bm \Psi_{k,i+1}|-z),0) \odot \text{sign}(\bm \Psi_{k,i+1})$ which costs $4L$ additions per each iteration at every node, where the comparisons required are counted as additions. Also, the 'prox. with $l_0$' algorithm requires an extra memory of size~$L$ for storing~$\bm \Psi_{k,i+1}$ at node~$k$.

\section{Conclusion}
We have derived the D-NLMM algorithm over distributed networks by applying the MH function, which is robust against impulsive noise. Furthermore, considering the underlying sparsity for parameters of interest, the D-SNLMM algorithm was proposed by incorporating the $l_0$-norm based zero-attractor into the D-NLMM recursion. The mean and mean-square behaviors of the proposed algorithms including the transient and steady-state were analyzed. This analysis does not require the integrals on the score function and the Price's theorem, which is more simple than the conventional approach for dealing with the score function. The theoretical expression has shown that D-SNLMM outperforms D-NLMM in sparse systems. Simulations in various noise environments have shown the advantages of our proposed algorithms over some existing diffusion algorithms. The theoretical models are also supported by simulations.

\appendices
\numberwithin{equation}{section}
\section{Proof of Theorem 2 }
\renewcommand{\theequation}{\thesection.\arabic{equation}}
Recalling that the elements in $f(\bm w_{k,i})$ given by~\eqref{017} are bounded and usually $\beta$ is a small positive number, we may neglect the effect of some terms associated with $\beta f(\bm w_{k,i})$ on the stability condition of the D-SNLMM algorithm (see~\cite{di2013sparse,liu2012diffusion,liu2014distributed,ShravanPoly2018} for similar paradigms). In other words, the bounds on step sizes of the D-SNLMM are the same as those of the D-NLMM algorithm. Therefore, we will deduce the convergence condition in the mean-square from the D-NLMM algorithm for simplicity.

By resorting to the squared Euclidean norms of both sides of~\eqref{021}, setting $\beta=0$, and taking the expectation, we arrive at:
\begin{equation}
\begin{array}{rcl}
\begin{aligned}
\label{0A1}
\text{E}\{||\widetilde{\bm \psi}_{k,i+1}||_2^2\} =& \text{E}\{||\widetilde{\bm w}_{k,i}||_{\bm \varSigma_k}^2\} + \\
&\mu_k^2 P_{u,k}^2(i) \sigma_{\theta,k}^2 \text{E} \left\lbrace \frac{1}{||\bm u_{k,i}||_2^2 } \right\rbrace,
\end{aligned}
\end{array}
\end{equation}
where $\text{E}\{||\widetilde{\bm w}_{k,i}||_{\bm \varSigma_k}^2\} \triangleq \text{E}\{\widetilde{\bm w}_{k,i}^\text{T} \bm \varSigma_k \widetilde{\bm w}_{k,i}\}$,
\begin{equation}
\begin{array}{rcl}
\begin{aligned}
\label{0A2}
\bm \varSigma_k = \bm I_L- \mu_k P_{u,k}(i) \left( 2 - \mu_k P_{u,k}(i)\right) \text{E}\{\bm A_{k,i}\}.
\end{aligned}
\end{array}
\end{equation}

Recalling that $\text{E} \{\bm A_{k,i}\}$ is the normalized covariance matrix of input regressor at node $k$, so it is symmetric and nonnegative definite. Thus, using the eigenvalue decomposition of $\text{E} \{\bm A_{k,i}\}$, namely, $\text{E} \{\bm A_{k,i}\} = \bm U_k \bm \varLambda_k \bm U_k^\text{T}$ with $\bm \varLambda_k =\text{diag}\{\lambda_{k,1},...,\lambda_{k,L}\}$ collecting the eigenvalues and $\bm U_k$ being the corresponding orthogonal matrix, then we rewrite~\eqref{0A1} and~\eqref{0A2} as follows:
\begin{equation}
\begin{array}{rcl}
\begin{aligned}
\label{0A3}
\text{E}\{||\widetilde{\bm \psi}_{k,i+1}||_2^2\}& = \text{E}\{||\widetilde{\bm w}_{k,i}||_{\bm \varSigma_{\varLambda,k}}^2\} + \\
&\mu_k^2 P_{u,k}^2(i) \sigma_{\theta,k}^2 \text{E} \left\lbrace \frac{1}{||\bm u_{k,i}||_2^2 } \right\rbrace,
\end{aligned}
\end{array}
\end{equation}
where
\begin{equation}
\begin{array}{rcl}
\begin{aligned}
\label{0A4}
\bm \varSigma_{\varLambda,k} = \bm I_L - \mu_k P_{u,k}(i) \left( 2 - \mu_k P_{u,k}(i)\right)  \bm \varLambda_k
\end{aligned}
\end{array}
\end{equation}
is also a diagonal matrix. If~\eqref{052} holds, i.e., $(2-\mu_k P_{u,k}(i))>0$, we can use~\eqref{0A4} to obtain the following inequality:
\begin{equation}
\begin{array}{rcl}
\begin{aligned}
\label{0A5}
\delta_{k,\min}(i) \text{E}\{||\widetilde{\bm w}_{k,i}||_2^2\}  &\leq \text{E}\{||\widetilde{\bm w}_{k,i}||_{\bm \varSigma_{\varLambda,k}}^2\} \\
&\leq \delta_{k,\max}(i) \text{E}\{||\widetilde{\bm w}_{k,i}||_2^2\},
\end{aligned}
\end{array}
\end{equation}
where
\begin{equation}
\begin{array}{rcl}
\begin{aligned}
\label{0A6}
\delta_{k,\min}(i) &= 1 - \mu_k P_{u,k}(i) \left( 2 - \mu_k P_{u,k}(i)\right) \lambda_{\max} (\text{E} \{\bm A_{k,i}\}), \\
\delta_{k,\max}(i) &= 1 - \mu_k P_{u,k}(i) \left( 2 - \mu_k P_{u,k}(i)\right) \lambda_{\min} (\text{E} \{\bm A_{k,i}\}).
\end{aligned}
\end{array}
\end{equation}

In view of \eqref{020b} being a linear convex combination of $\{\bm \psi_{m,i+1}\}$, by taking advantage of Jensen's inequality \cite{chen2012diffusion,chen2013distributed}, the following inequality holds
\begin{equation}
\begin{array}{rcl}
\begin{aligned}
\label{0A7}
\text{E}\{||\widetilde{\bm w}_{k,i+1}||_2^2\} \leq \sum_{m\in\mathcal{N}_k} c_{m,k} \text{E}\{||\widetilde{\bm \psi}_{k,i+1}||_2^2\}.
\end{aligned}
\end{array}
\end{equation}

Introducing the following vectors:
\begin{equation}
\begin{array}{rcl}
\begin{aligned}
\label{0A8}
\mathcal{\bm X}_i & \triangleq \text{col} \left\lbrace \text{E}\{||\widetilde{\bm \psi}_{1,i}||_2^2\}, ..., \text{E}\{||\widetilde{\bm \psi}_{N,i}||_2^2\} \right\rbrace  \\
\mathcal{\bm Y}_i & \triangleq \text{col} \left\lbrace \text{E}\{||\widetilde{\bm w}_{1,i}||_2^2\}, ..., \text{E}\{||\widetilde{\bm w}_{N,i}||_2^2\} \right\rbrace  \\
\mathcal{\bm Z}_i & \triangleq \text{col}\left\lbrace \mu_1^2 P_{u,1}^2(i) \sigma_{\theta,1}^2 \text{E} \left\lbrace \frac{1}{||\bm u_{1,i}||_2^2 } \right\rbrace, ...,\right. \\
&\;\;\;\;\;\;\;\;\;\;\;\;\left. \mu_N^2 P_{u,N}^2(i) \sigma_{\theta,N}^2 \text{E} \left\lbrace \frac{1}{||\bm u_{N,i}||_2^2 } \right\rbrace  \right\rbrace,  \\
\end{aligned}
\end{array}
\end{equation}
and the diagonal matrix
\begin{equation}
\begin{array}{rcl}
\begin{aligned}
\label{0A9}
\bm \delta_i  & \triangleq \text{diag} \left\lbrace \delta_{1,\max}(i), ..., \delta_{N,\max}(i) \right\rbrace,  \\
\end{aligned}
\end{array}
\end{equation}
then gathering \eqref{0A3}, \eqref{0A5}, and \eqref{0A7} at all nodes implies that
\begin{subequations} \label{0A10}
    \begin{align}
    \mathcal{\bm X}_{i+1} &\preceq \bm \delta_i \mathcal{\bm Y}_{i} + \mathcal{\bm Z}_i, \label{0A10a}\\
    \mathcal{\bm Y}_{i+1} &\preceq \bm C^\text{T} \mathcal{\bm X}_{i+1}, \label{0A10b}
    \end{align}
\end{subequations}
where $\preceq$ denotes the entry-wise comparison. Since the entries of the matrix $\bm C$ are nonnegative, \eqref{0A10a} and \eqref{0A10b} further lead to
\begin{equation}
\begin{array}{rcl}
\begin{aligned}
\label{0A11}
\mathcal{\bm Y}_{i+1} \preceq \bm C^\text{T} \bm \delta_i \mathcal{\bm Y}_{i}  + \bm C^\text{T} \mathcal{\bm Z}_i.
\end{aligned}
\end{array}
\end{equation}

Taking the $\infty$-norm of both sides of \eqref{0A11} and again using $||\bm C^\text{T}||_\infty=1 $, we obtain
\begin{equation}
\begin{array}{rcl}
\begin{aligned}
\label{0A12}
||\mathcal{\bm Y}_{i+1}||_\infty &\leq ||\bm C^\text{T}||_\infty \cdot ||\bm \delta_i||_\infty \cdot ||\mathcal{\bm Y}_{i}||_\infty + ||\bm C^\text{T}||_\infty \cdot ||\mathcal{\bm Z}_i||_\infty\\
&= ||\bm \delta_i||_\infty \cdot ||\mathcal{\bm Y}_{i}||_\infty + ||\mathcal{\bm Z}_i||_\infty.\\
\end{aligned}
\end{array}
\end{equation}

Given that $||\mathcal{\bm Z}_i||_\infty = \max \limits_{1\leq k \leq N} \mu_k^2 P_{u,k}^2(i) \sigma_{\theta,k}^2 \text{E} \left\lbrace \frac{1}{||\bm u_{k,i}||_2^2 } \right\rbrace$ is bounded,~\eqref{0A12} converges if and only if $||\bm \delta_i||_\infty<1$, which is equivalent to requiring
\begin{equation}
\begin{array}{rcl}
\begin{aligned}
\label{0A13}
|1 - \mu_k P_{u,k}(i) \left( 2 - \mu_k P_{u,k}(i)\right) \lambda_{\min} (\text{E} \{\bm A_{k,i}\})| < 1
\end{aligned}
\end{array}
\end{equation}
for $k=1,...,N$. From~\eqref{0A13}, the inequality~\eqref{052} is acquired.

\section{Proximal Variants of The D-SNLMM Algorithm}
In Section III.~B, we have derived the D-SNLMM algorithm for solving the distributed estimation problem~\eqref{012} by virtue of the known SGD rule. In this appendix, we will show briefly how to deduce the distributed proximal solution. Accordingly, we recast~\eqref{012} as
\begin{equation}
\label{0B1}
\begin{array}{rcl}
\begin{aligned}
&\min \limits_{\bm w_k} J_k^{loc}(i),\\
&J_k^{loc}(i) = G_k(i) + \beta F(\bm w_k),
\end{aligned}
\end{array}
\end{equation}
where
\begin{equation}
\label{0B2}
\begin{array}{rcl}
\begin{aligned}
G_k(i) = \sum\limits_{m\in \mathcal{N}_k} c_{m,k} g_m^{-1} \text{E} \left\lbrace \varphi (d_m(i)-\bm u_{m,i}^\text{T}\bm w_k) \right\rbrace.
\end{aligned}
\end{array}
\end{equation}

The forward-backward splitting method~\cite{chen2014dictionary,nassif2016proximal} to get the minimizer of~\eqref{0B1} consists of two steps. It firstly performs the forward step to minimize $G_k(i)$ according to the SGD rule (also see the derivation in Section III.~A):
\begin{subequations}
    \label{0B3}
    \begin{align}
    \bm \psi_{k,i+1} &= \bm w_{k,i} + \mu_k g_k^{-1} \bm u_{k,i} \varphi' (e_k(i)), \label{0B3a}\\
    \bm \Psi_{k,i+1} &= \sum\limits_{m\in \mathcal{N}_k} c_{m,k} \bm \psi_{m,i+1}. \label{0B3b}
    \end{align}
\end{subequations}
where $g_k=1$ and $g_k=\left\| \bm u_{k,i}\right\|_2^2$ correspond to the non-normalized and normalized types, respectively. Subsequently, it performs the proximal step:
\begin{equation}
\label{0B4}
\begin{array}{rcl}
\begin{aligned}
\bm w_{k,i+1} = \text{prox}_{\mu_k\beta F}(\bm \Psi_{k,i+1}),
\end{aligned}
\end{array}
\end{equation}
where $\text{prox}(\cdot)$ is referred to as the proximal operator of index $\mu_k\beta\in (0, +\infty)$, defined by
\begin{equation}
\label{0B5}
\begin{array}{rcl}
\begin{aligned}
\text{prox}_{\mu_k\beta F}(\bm \Psi_{k,i+1}) = \min \limits_{\bm w_k}\beta F(\bm w_k) + \frac{1}{2\mu_k}||\bm \Psi_{k,i+1} - \bm w_k||_2^2,
\end{aligned}
\end{array}
\end{equation}
For ease of implementation, a closed-form solution of~\eqref{0B5} is necessary. Benefited from the subdifferential feature of the function~\cite{parikh2014proximal}, we can solve~\eqref{0B5} to
\begin{equation}
\label{0B6}
\begin{array}{rcl}
\begin{aligned}
\bm 0 \in \beta \partial F(\bm w_k) - \frac{1}{2\mu_k}  (\bm \Psi_{k,i+1} - \bm w_k).
\end{aligned}
\end{array}
\end{equation}

If the sparse regularization $F(\bm w_k)=\sum \limits_{l=1}^L |[\bm w]_l|$ (the $l_1$-norm in Table~\ref{table_1}) is used, then~\eqref{0B6} will become
\begin{equation}
\label{0B7}
\begin{array}{rcl}
\begin{aligned}
\bm 0 \in \beta \text{sign}(\bm w_k) - \frac{1}{2\mu_k}  (\bm \Psi_{k,i+1} - \bm w_k),
\end{aligned}
\end{array}
\end{equation}
which leads to a popular proximal operator, called the soft-threshold~\cite{wee2013proximal}:
\begin{equation}
\label{0B8}
\begin{array}{rcl}
\begin{aligned}
\text{prox}_{\mu_k\beta ||\cdot||_1}(\bm \Psi_{k,i+1}) =& \max(|\bm \Psi_{k,i+1}|-\mu_k\beta,0) \odot \\
& \text{sign}(\bm \Psi_{k,i+1}),
\end{aligned}
\end{array}
\end{equation}
where $\odot$ denotes the element-wise product of two vectors. To significantly utilize the sparsity, we still consider the $l_0$-norm approximation in~\eqref{015} for $F(\bm w_k)$ so that
\begin{equation}
\label{0B9}
\begin{array}{rcl}
\begin{aligned}
\bm 0 \in \beta f_w(\bm w_k) \odot \text{sign}(\bm w_k) - \frac{1}{2\mu_k}  (\bm \Psi_{k,i+1} - \bm w_k),
\end{aligned}
\end{array}
\end{equation}
where the column vector $f_w(\bm w_k)$ consists of $L$ entries
\begin{equation}
\label{0B10}
\begin{array}{rcl}
\begin{aligned}
f_w([\bm w]_l) = \upsilon \text{exp}^{-\upsilon|[\bm w]_l|},\;l=1,...,L.
\end{aligned}
\end{array}
\end{equation}
It is clear that~\eqref{0B9} can be considered as a weighted variant of~\eqref{0B7}, and then the associated proximal step with the $l_0$-norm is represented by the weighted soft-threshold below:
\begin{equation}
\label{0B11}
\begin{array}{rcl}
\begin{aligned}
\text{prox}_{\mu_k\beta ||\cdot||_0}(\bm \Psi_{k,i+1}) =& \max(|\bm \Psi_{k,i+1}|-\mu_k\beta f_w(\bm \Psi_{k,i+1}),0) \odot \\
& \text{sign}(\bm \Psi_{k,i+1}).
\end{aligned}
\end{array}
\end{equation}

In the algorithm's implementation, we also employ the first order Taylor series of the exponential function which helps to reduce the complexity of~\eqref{0B11}, namely,
\begin{equation}
\label{0B12}
f_\Psi([\bm \Psi]_l)=\left\{ \begin{aligned}
&\upsilon^2 [\bm \Psi]_l + \upsilon, \text{ if } -\frac{1}{\upsilon}\leq [\bm \Psi]_l<0\\
&-\upsilon^2 [\bm \Psi]_l + \upsilon, \text{ if } 0<[\bm \Psi]_l \leq \frac{1}{\upsilon} \\
&0, \text{elsewhere}.
\end{aligned} \right.
\end{equation}

Note that other sparse regularization criteria in Table~\ref{table_1} can also be extended under this weighted proximal formalism. Moreover, these distributed proximal algorithms have also the same stability condition~\eqref{064} as the D-NLMM algorithm, with the aid of references~\cite{di2013sparse,wee2013proximal}. However, the detailed proof is omitted due to lack of space. In fact, because of involving the proximal step~\eqref{0B11}, the theoretical analysis of the proximal algorithm will also be more complicated than that of the D-SNLMM algorithm.

\ifCLASSOPTIONcaptionsoff
  \newpage
\fi

\bibliographystyle{IEEEtran}
\bibliography{IEEEabrv,mybibfile}

\end{document}